\documentclass[10pt,twocolumn,letterpaper]{article}

\usepackage{cvpr}              
\usepackage{algorithm}
\usepackage{algorithmic}
\usepackage{amsmath}
\usepackage{multirow}
\usepackage[normalem]{ulem}
\useunder{\uline}{\ul}{}
\usepackage{booktabs}
\usepackage{pifont}
\usepackage{adjustbox}
\usepackage{setspace}
\usepackage{multirow}
\usepackage{amsmath}
\usepackage{amssymb}
\usepackage{booktabs}
\usepackage{enumitem}
\usepackage{colortbl}

\definecolor{cvprblue}{rgb}{0.21,0.49,0.74}
\usepackage[pagebackref,breaklinks,colorlinks,allcolors=cvprblue]{hyperref}

\def\paperID{30053} 
\def\confName{CVPR}
\def\confYear{2026}

\title{\textit{ViRectify}: A Challenging Benchmark for Video Reasoning Correction with Multimodal Large Language Models}

\author{Xusen Hei$^*$, Jiali Chen$^*$, Jinyu Yang, Mengchen Zhao, Yi Cai \\
South China University of Technology \\
{\tt\small \{sexusenhei, segarychen\}@mail.scut.edu.cn} \\
}

\begin{document}
\maketitle
\begin{abstract}
As multimodal large language models (MLLMs) frequently exhibit errors in complex video reasoning scenarios, correcting these errors is critical for uncovering their weaknesses and improving performance.
However, existing 
benchmarks lack
systematic evaluation of MLLMs' ability to identify and correct these video reasoning errors.
To bridge this gap, we propose \textit{ViRectify}, a comprehensive benchmark 
to evaluate their fine-grained correction capability.
Through an AI-assisted annotation pipeline with human verification, we construct a dataset of over 30\textit{K} instances spanning dynamic perception, scientific reasoning, and embodied decision-making domains. 
In \textit{ViRectify}, we challenge MLLMs to perform step-wise error identification and generate rationales with key video evidence grounding.
In addition, we further propose the trajectory evidence-driven correction framework, comprising step-wise error trajectory and reward modeling on visual evidence-grounded correction. 
It encourages the model to explicitly concentrate on error propagation and key timestamps for correction.
Extensive evaluation across 16 advanced MLLMs demonstrates that our \textit{ViRectify} serves as a challenging testbed, where GPT-5 achieves only 31.94\% correction accuracy.
Our framework enables a Qwen2.5-VL-7B to consistently outperform the variants of 72B on \textit{ViRectify}, showing the effectiveness of our approach. 
Further analysis uncovers systematic asymmetries in error correction across models, and our dataset is also a valuable data resource to perform reflection learning.
We believe \textit{ViRectify} provides a new direction for comprehensively evaluating the advanced MLLMs in video reasoning\footnote{Our dataset and code will be released at the project page: \url{https://github.com/black1948/ViRectify}}. 
\end{abstract}
\section{Introduction}
\begin{figure}[!]
  \centering
  \includegraphics[scale=0.478]{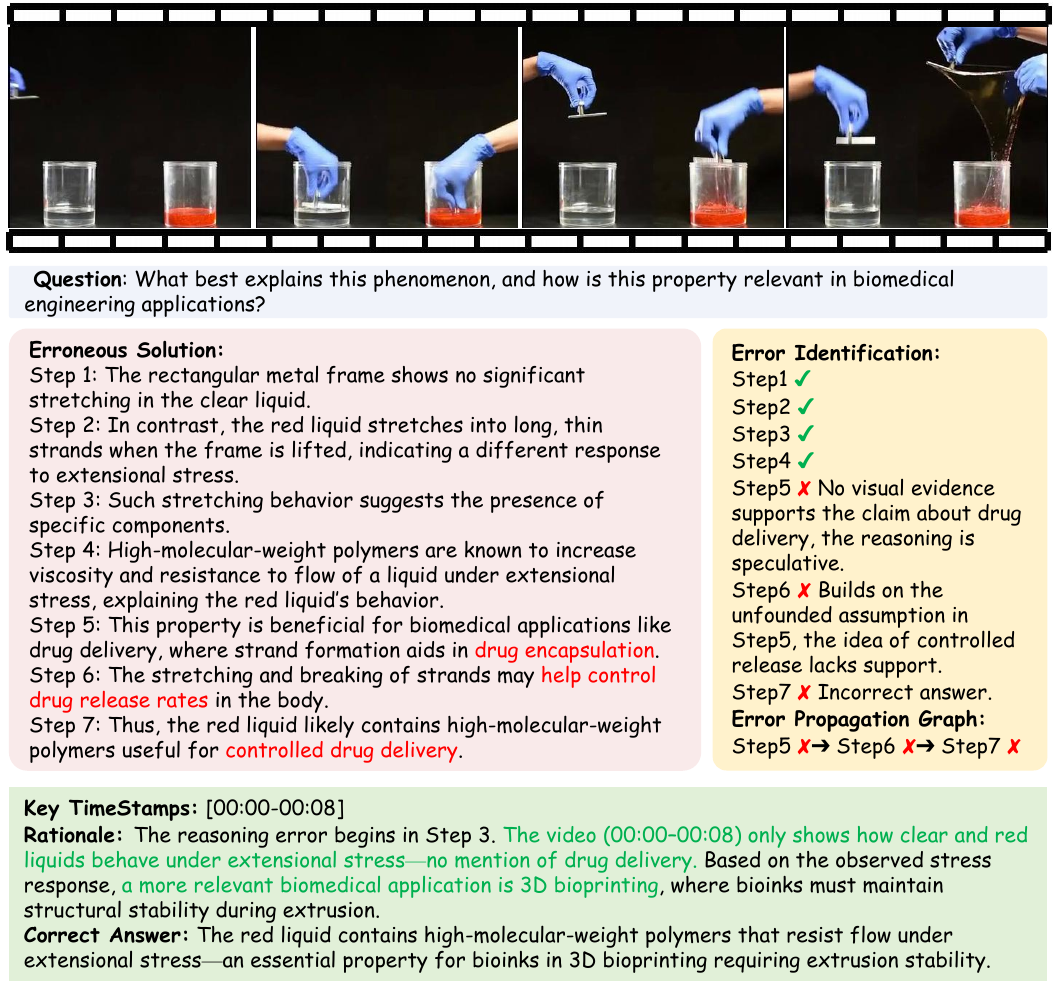}
  \caption{An example from our \textit{ViRectify} benchmark. Green denotes reasoning consistent with the video evidence. Red marks steps that are incorrect or unsupported.}
\label{pic_intro_case}
\end{figure}
Current multimodal large language models (MLLMs) have achieved significant progress in video understanding \cite{stream, streamvllm, LVAgent, TOPA, VCA, Flash-VStream} and reasoning \cite{mmvu, motionbench,intentqa, MECD} tasks. Typically, MLLMs adopt the chain-of-thought (CoT) paradigm \cite{cot, selfcot}, where the model generates intermediate reasoning steps before reaching the final answer. However, the reasoning chains produced by these models may contain errors, raising concerns about their reliability and highlighting the need for correction mechanisms in video reasoning. In real-world educational settings, human teachers are expected to identify and correct reasoning errors. However, enabling models to emulate this diagnostic and corrective process remains underexplored.

Unlike prior tasks that primarily focus on evaluating the correctness of final answers \cite{mmvu, mmworld, LVBench}, our task targets fine-grained correction of the reasoning process. Specifically, we require MLLMs to identify erroneous steps within the reasoning chain, analyze the propagation of these errors across subsequent steps, and generate explanatory rationales based on video evidence. Existing benchmarks \cite{videoespresso} mainly focus on generating correct reasoning chains, with little emphasis on identifying and correcting erroneous reasoning chains.
While prior works \cite{projudge, visco} explore step-level error identification in static images, we extend this to the more challenging video domain and additionally examine the propagation of errors throughout the reasoning process, as shown in Fig. \ref{pic_intro_case}.
Despite recent efforts, there remains a lack of comprehensive benchmarks specifically designed to evaluate correction capabilities in video reasoning tasks.

To address this gap, we introduce \textit{ViRectify}, a comprehensive benchmark for systematically evaluating the ability of models to identify and correct errors in video reasoning tasks. Our benchmark exhibits the following four key characteristics:
\textbf{(i) Broad Domain Coverage:} The benchmark encompasses three key video reasoning domains (\ie, dynamic perception, scientific reasoning, and embodied decision-making) across seven diverse datasets, ensuring strong generalization and applicability.
\textbf{(ii) Realistic and Diverse Error Patterns:} We collect over 30,000 annotated samples through an automated pipeline, combining synthetic errors from a proprietary model and natural errors from 8 MLLMs with varying sizes and architectures, capturing a broad spectrum of real-world reasoning failures.
\textbf{(iii) Fine-Grained Error Identification and Tracing:} Each step in erroneous solutions is labeled with correctness and accompanied by concise explanations for incorrect steps. We further construct error propagation graphs, enabling evaluation of a model's ability to identify and trace root causes of reasoning failures.
\textbf{(iv) Error Correction Grounded with Video Evidence:} Using GPT-o4-mini, we generate corrections based on video evidence, allowing assessment of a model's ability to revise errors with context-aware rationales.

We evaluate 16 multimodal large language models (MLLMs) on \textit{ViRectify}.
Our results show that proprietary models notably outperform open-source ones in error identification and correction, yet even the strongest models achieve only around 30\% correction accuracy, underscoring a major gap in current capabilities.
To bridge this gap, we propose a trajectory evidence-driven correction framework with a two-stage training paradigm. The first stage employs a graph-based loss to guide fine-grained reasoning over error trajectories, enhancing step-wise error identification. The second stage leverages GRPO \cite{deepseekmath} with a temporal alignment reward to improve the model’s use of key video segments for evidence-driven correction.
Our framework empowers Qwen2.5-VL-7B to outperform proprietary models like GPT-5 in error identification and achieve comparable correction accuracy, highlighting its effectiveness in narrowing the performance gap.
To further examine model behavior, we conducted analytical experiments across error types and found logical errors to be the most challenging for both identification and correction. Models with stronger reasoning abilities show higher correction accuracy in this category, and our framework yields notable gains. Moreover, our dataset supports reflection learning, enhancing error-aware and video-grounded reasoning.

Our main contributions are summarized as follows:
(i) We introduce \textit{ViRectify}, a systematic and large-scale benchmark designed to rigorously assess models' capabilities in detecting and correcting reasoning errors in video reasoning tasks.
(ii) We conduct comprehensive experiments on \textit{ViRectify}, revealing notable limitations of current MLLMs in accurate error identification and correction. To address these issues, we propose a trajectory evidence-driven correction framework that significantly improves model performance.
(iii) Further analysis shows persistent correction bottlenecks and systematic asymmetries across models. \textit{ViRectify} provides a valuable resource for facilitating reflection and fostering more reasoning-aware model behavior.
\section{Related Work}
\label{sec:realted_work}
\subsection{Multimodal Large Language Models}
Foundational MLLMs (\eg, InstructBLIP \cite{instructblip} and LLaVA \cite{llava}) adopt multimodal instruction tuning as the core paradigm to enable general text-image understanding.
MLLMs like GPT-o3 \cite{gpt-o3}, Gemini-2.5 \cite{gemini2.5}, and Seed-VL \cite{seed-vl}, as well as open-source counterparts (\eg, MiMo-VL \cite{mimo-vl} and GLM-4.1V-Thinking \cite{glm4.1v}) utilize verifiable RL rewards to improve complex reasoning ability.
Additionally, video-centric MLLMs including Video-LLaMA \cite{video-llama} and LLaVA-Next \cite{llava-next} introduce spatial-temporal aggregation mechanisms.
Recent studies \cite{VideoR1, VideoChat-R1, VideoRFT} have incorporated reinforcement learning to strengthen models' video reasoning capability.


\subsection{Video Understanding and Reasoning}
Existing VideoQA datasets can be broadly categorized into three types based on their focus and complexity. At the perceptual level, tasks focus on visual attribute recognition, spatial relations, and temporal boundaries \cite{video-mme, intentqa}.
Reasoning-oriented benchmarks further increase task complexity \cite{mmworld, mmvu, video-mmmu}, requiring procedural, causal, and domain-specific reasoning. 
Embodied video tasks \cite{egotextvqa, hd-epic} challenge models with egocentric views and interactive physical reasoning.
Existing benchmarks emphasize outcome correctness but often neglect the evaluation of a model’s ability to identify and correct reasoning errors, a key aspect of robust video reasoning. To address this gap, we introduce \textit{ViRectify} for correction on video reasoning.

\section{\textit{ViRectify} Benchmark}

\begin{figure*}[!]
  \centering
  \includegraphics[scale=0.39]{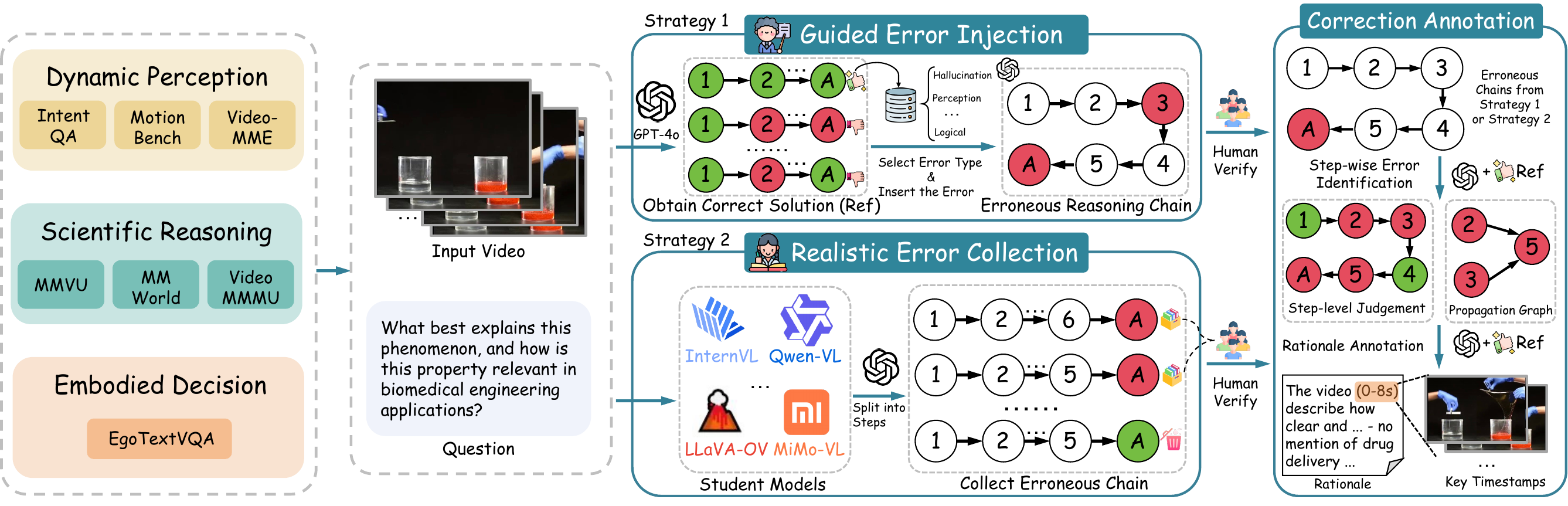}
  \caption{Construction process of \textit{ViRectify}. We adopt two strategies to collect erroneous solutions: (1) injecting errors with a proprietary model, and (2) collecting naturally occurring reasoning mistakes from smaller models. 
  Then we annotate the erroneous reasoning chains with step-level error identifications and rationales grounded in video evidence.
  }
\label{data_anno}
\end{figure*}

In this section, we introduce the construction process of our \textit{ViRectify} benchmark, as illustrated in Fig. \ref{data_anno}.
Specifically, 
\textit{ViRectify} consists of tuples $(x, s, c)$, where $x$ represents the input video-question pair, $\mathbf{s}_e=[s_1,s_2,...,s_N]$ with $N$ reasoning steps is the step-wise errorneous solution, 
and the error correction component $c$ conducts step-wise error identification with concise explanations, models error propagation across reasoning steps, integrates critical video evidence, and produces a video-grounded rationale to guide correction toward the accurate solution.
In the following subsections, we first describe the data sources of \textit{ViRectify} across three representative video domains.
Then, we elaborate on our AI-assisted annotation pipeline, which involves automatic erroneous reasoning creation and correction labeling with human verification.
Finally, we provide comprehensive statistics and analysis of our benchmark.

\subsection{Dataset Source} \label{sec:data_source}
To comprehensively assess the performance of MLLMs across diverse scenarios,
\textit{ViRectify} comprises datasets from three domains: 
dynamic perception, scientific reasoning, and embodied decision-making.
And we apply difficulty-based sampling (see Appendix) followed by manual verification to ensure high-quality, challenging questions.

\noindent \textbf{Dynamic Perception.} 
It challenges MLLMs to perform fine-grained perception tasks in dynamic videos, including object recognition, motion analysis and inference of implicit intentions.
MotionBench \cite{motionbench} evaluates object recognition and basic motion perception using synthetic and real-world videos.
Video-MME \cite{video-mme} contains long open-domain videos with dense visual and textual cues, challenging models on multi-object perception and temporal understanding.
IntentQA \cite{intentqa} focuses on inferring implicit human intentions from short video interactions.
We obtain 2,123 challenging examples from IntentQA, 283 from Video-MME, and 574 from MotionBench.

\noindent \textbf{Scientific Reasoning.}
It focuses on knowledge-intensive reasoning across multiple scientific disciplines (\eg, physics, chemistry and medicine).
We choose publicly available datasets (\ie, MMVU \cite{mmvu}, MMWorld \cite{mmworld}, and Video-MMMU \cite{video-mmmu}) to cover a wide range of scientific video reasoning challenges. 
Specifically, MMVU and Video-MMMU focus on questions requiring domain-specific scientific knowledge, while MMWorld centers on tasks involving world modeling and counterfactual reasoning.
We adopt 595 samples from MMVU, 1,010 from MMWorld, and 279 from Video-MMMU.

\noindent \textbf{Embodied Decision-making.}
Extending from prior reasoning tasks, this domain addresses real-world video reasoning applications by assessing models’ competence in decision-making within complex egocentric settings.
We incorporate EgoTextVQA \cite{egotextvqa}, an egocentric video question answering dataset centered on real-life scenarios such as driving and home environments. 
We select 1,224 high-quality, challenging samples from EgoTextVQA.

\subsection{Automatic Erroneous Reasoning Creation}
We collect erroneous solutions on the high-quality, challenging examples curated through manual verification.
To ensure that \textit{ViRectify} offers a diverse and sufficiently challenging set of erroneous solutions for correction.
We adopt two complementary strategies to collect these solutions: (i) injecting errors into correct solutions with a powerful proprietary GPT-4o \cite{gpt4} to obtain plausible yet incorrect reasoning chains, and (ii) leveraging small MLLMs to simulate student-like mistakes arising from natural reasoning errors. 

\noindent \textbf{Error Injection with Proprietary Model.}
As our task demands temporally dense and frame-level visual understanding, existing proprietary models are constrained by their limited ability to process multiple consecutive frames.
To address this limitation, we employ the open-source Qwen2.5-VL-32B \cite{qwen2.5-vl} to perform question-guided caption to capture the fine-grained semantic information of the video. 
We segment videos into 1-second clips and sample frames at 30 FPS. These frames, along with the question as context, are input to Qwen2.5-VL-32B for caption generation. Semantically similar adjacent captions are then merged into temporally continuous segments, each annotated with start and end timestamps (\eg, 10s–13s) and describing a coherent video event.
Detailed settings are provided in Appendix. 
We use GPT-4o \cite{gpt4} to generate chain-of-thought (CoT) solutions based on video segment captions and their corresponding questions. To ensure reliable reasoning chains for subsequent error injection, we retain only those whose final answers match the reference answers, verified by explicit comparison. 
One of four predefined error types (\ie, question misunderstanding, hallucination, reasoning, or visual perception) is systematically injected into each correct solution, resulting in 7,788 erroneous solutions over 6,088 video–question pairs.

\noindent \textbf{Realistic Error Collection from Small MLLMs.}
To emulate the reasoning failures frequently made by human students in real-world scenarios, we select small MLLMs with varying sizes (from 3B to 8B) (\ie, Qwen2.5-VL-3/7B \cite{qwen2.5-vl}, InternVL3-8B \cite{internvl3}, LLaVA-OneVision \cite{llava-ov}, R1-Onevision-7B-RL \cite{r1-onevision}, ViGal-7B \cite{vigal}, MM-Eureka-7B \cite{mm-eureka} and MiMo-VL-7B-RL \cite{mimo-vl}).
We prompt small models to generate reasoning chains for video–question pairs with verified correct answers.
To ensure consistent step granularity, the resulting solutions are split into standardized steps for downstream correction annotation. The chains are then checked against the reference answers, and only those with incorrect predictions are retained. These are further categorized into four predefined error types: question misunderstanding, hallucination, reasoning, and visual perception. In total, the procedure yields 26,130 erroneous solutions over 6,088 video–question pairs.

\noindent \textbf{Human Verification and Duplication Filtering.}
To ensure that the collected erroneous solutions are both diverse and sufficiently challenging for correction, we invite 3 human expert and implement a three-stage human verification process. 
Details regarding the human experts involved in the verification process are provided in Appendix.
Specifically, the process consists of:
(i) \textbf{Correctness Validation:}
Beyond answer-based filtering, experts reviewed the full reasoning chains, confirming that over 98\% demonstrate logical coherence. 
It validates the filtering approach and ensures high-quality reasoning chains for error injection.
(ii) \textbf{Error Type Validation:}
Verify that annotated error types align with the actual reasoning flaws present in the solutions.
(iii) \textbf{Quality Filtering:}
Remove solutions that are internally inconsistent, illogical, or self-contradictory.
Finally, we merge data from both error collection methods and deduplicate by retaining only one instance per video-question pair when erroneous solutions are identical or semantically similar.
In total, we curate over 30\textit{K} erroneous solutions, each undergoing thorough human verification and deduplication to ensure quality and uniqueness.
\begin{minipage}[t]{0.22\textwidth}
  \begin{table}[H]
    \centering
    \vspace{0pt}
    \resizebox{\linewidth}{!}{
      \begin{tabular}{l|c}
        \toprule[1pt]
        \textbf{Statistics}              & \textbf{Number} \\ \midrule
        Error Type                       &                 \\
        - Question Misunderstanding      & 5,242           \\
        - Hallucination                  & 2,915           \\
        - Logical                        & 22,071          \\
        - Visual Perception              & 3,563           \\ \midrule
        Error Steps                      &                 \\
        - 1$\sim$3 steps                 & 12,738          \\
        - 4$\sim$7 steps                 & 15,900          \\
        - $>$ 7 steps                    & 5,278           \\ \midrule
        Key Timestamp length             &                 \\
        - 0$\sim$10s                     & 10,002          \\
        - 10$\sim$20s                    & 17,142          \\
        - $>$ 20s                        & 6,774           \\ \midrule[1pt]
      \end{tabular}
    }
    \caption{Key statistics of our \textit{ViRectify}.}
    \label{stats}
  \end{table}
\end{minipage}%
\hfill
\begin{minipage}[t]{0.22\textwidth}
  \begin{figure}[H]
    \centering
    \vspace{2pt}
    \includegraphics[width=\linewidth]{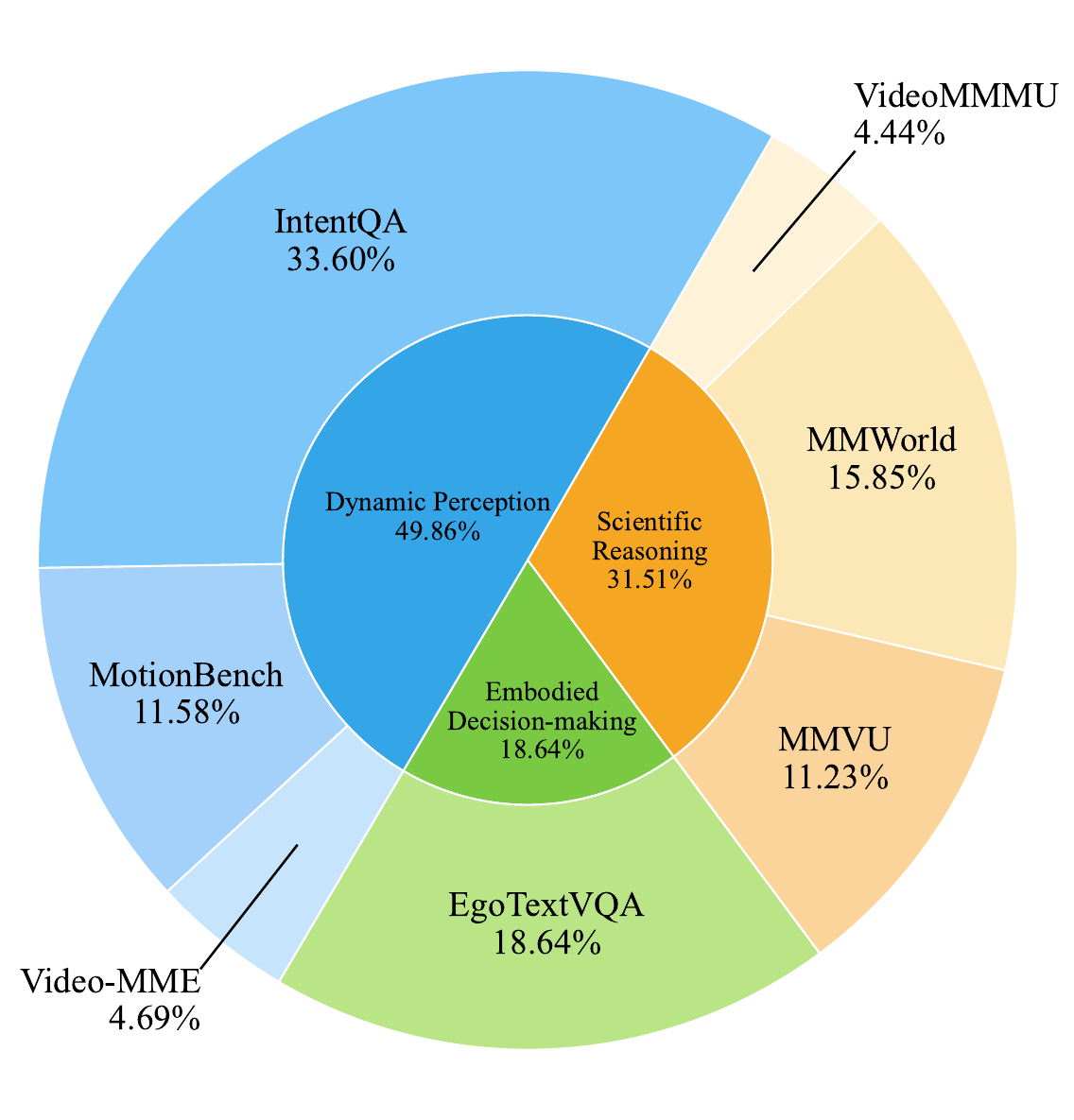}
    \caption{Domain distribution of \textit{ViRectify}.}
    \label{domain}
  \end{figure}
\end{minipage}
\subsection{Error Correction Annotation}
We employ GPT-o4-mini \cite{gpt-o3} to annotate corrections for the collected erroneous solutions by identifying flawed steps, tracing error propagation, and generating rationales based on critical video evidence. To ensure annotation quality, a subset of the corrections is further verified by human experts, confirming the accuracy and completeness of the automated correction process (see Appendix).

\noindent \textbf{Error Identification Annotion.}
We perform step-wise error identification for each erroneous solution using the video caption, the question, and the correct verified solution as context. For each incorrect step, a brief explanation of the error was provided. The propagation of errors was traced by constructing an error propagation graph encoded as an adjacency list, where each entry (e.g., “step 3: [step 1]”) indicates that the error in a given step originates from an earlier step.

\noindent \textbf{Video-Grounded Rationale Annotation.}
Given the video captions with timestamps, the associated question, the original correct reasoning chain, the erroneous solution, and the detailed error identification, we perform fine-grained visual grounding by pinpointing critical timestamps that serve as essential evidence for correcting the identified errors. Based on the visual content at these key timestamps, a video-grounded rationale is generated to guide the revision of the reasoning process to arrive at the correct answer.

\subsection{Dataset Statistics}

We analyze the composition and characteristics of the \textit{ViRectify} benchmark, with detailed statistics presented in Table \ref{stats} and Fig. \ref{domain}. 
\textbf{Domain Coverage:} Our benchmark comprises three domains: Dynamic Perception, Scientific Reasoning, and Situated Decision-Making, containing 16,910, 10,686, and 6,322 samples respectively, accounting for 49.86\%, 31.51\%, and 18.64\% of the total. 
\textbf{Error Type Distribution:} 
Our benchmark includes four error categories: question misunderstanding, hallucination, reasoning, and visual perception. Logical errors dominate, with 22,071 instances accounting for over 60\% of the dataset, underscoring the central challenge of reasoning in video-based question answering. The remaining errors include 5,242 cases (15.51\%) of question misunderstanding, 2,915 (8.63\%) of hallucination, and 3,563 (10.54\%) of visual perception.
\textbf{Complexity of Error Chains:}
Erroneous solutions in our benchmark contain an average of 4.78 incorrect steps, reflecting the overall complexity of the correction task. Specifically, 12,738 samples (37.57\%) contain 3 or fewer errors, 15,900 (46.88\%) have 4–7 errors, and 5,278 (15.57\%) include more than 7, highlighting the challenge of identifying and correcting multiple reasoning flaws.
\textbf{Temporal Scope of Video Evidence:}
In our benchmark, 29.48\% (10,002 instances) of key timestamps span 0–10 seconds, 50.54\% (17,142) span 10–20 seconds, and 19.97\% (6,774) exceed 20 seconds. This distribution indicates that correcting flawed reasoning often relies on sustained visual evidence rather than brief cues, underscoring the importance of integrating extended visual context for accurate error identification and correction.
\section{Trajectory Evidence-Driven Correction}
Given a video-question pair and its corresponding erroneous solution, our goal is to identify the erroneous steps in the solution and correct them accordingly. 
To enable fine-grained error identification and visual-grounded correction, we propose a trajectory evidence-driven framework.
Our trajectory evidence-driven correction framework consists of two fine-tuning stages: step-wise error trajectory modeling via SFT, which supervises the model to identify errors at each step, and evidence-guided correction via RL, which guides the model to revise these errors based on visual cues. 

\subsection{Step-wise Error Trajectory Modeling via SFT}
In the initial stage, the model is supervised to identify and trace errors step by step, using annotated correctness labels to learn where and how reasoning deviates from the correct path, while gaining a preliminary ability to revise incorrect steps.
We split the \textit{ViRectify} benchmark into training and test sets with a 15:1 video-level ratio, ensuring no video overlap across sets. 
During training, we perform supervised fine-tuning (SFT) on the training set using a combination of the standard cross-entropy loss and a graph-based loss. The graph loss is defined over the error propagation graph, where each node corresponds to a reasoning step and edges indicate whether an error in one step affects another. Specifically, given the predicted graph ($\hat{G}$) and the ground-truth graph ($G$), the graph loss is computed as one minus the Graph Jaccard similarity:
\begin{equation}
  \mathcal{L}_{\text{graph}} = 1 - \frac{|E(G) \cap E(\hat{G})|}{|E(G) \cup E(\hat{G})|},  
\end{equation}
where $E(G)$ and $E(\hat{G})$ denote the sets of edges in the ground-truth and predicted graphs, respectively. This loss encourages the model to capture the structural dependencies among reasoning steps by promoting consistency between predicted and true error propagation relations.


\subsection{Evidence-guided Correction via RL}

While first-stage training achieves high error identification accuracy, the model still struggles with effective error correction. To address this, we adopt GRPO \cite{deepseekmath} for continued training, enhancing the model’s ability to leverage video information for accurate correction. 
Before training, we estimate the intrinsic difficulty of each sample by performing pass@k evaluations on the training set using the first-stage checkpoint, and select samples of moderate difficulty, specifically those with pass@k values between 1 and 4.
Following \cite{deepseek-r1}, we design a composite reward function, consisting of three components: temporal alignment, answer accuracy, and format consistency rewards.

\noindent \textbf{Temporal Alignment Reward.}
To enhance the model's understanding and utilization of critical video segments, we design a temporal alignment reward that measures the overlap between the predicted and reference timestamps:
\begin{equation}
    \mathcal{R}_{\text{time}} = \text{IoU}(\hat{T}, T^{*}),
\end{equation}
where $\hat{T}$ denotes the set of predicted temporal segments referenced in the correction, and $T^{*}$ is the set of annotated ground-truth segments.

\noindent \textbf{Answer Accuracy Reward.}
To evaluate whether the corrected reasoning produces a valid and contextually appropriate answer, we adopt GPT-4o \cite{gpt4} as an automatic evaluator to assess the semantic alignment between the predicted answer and the ground-truth reference. Specifically, a binary reward is assigned: a value of 1 is given if GPT-4o determines that the predicted answer is semantically equivalent to the reference answer and adequately addresses the question; otherwise, a value of 0 is assigned.

\noindent \textbf{Format Consistency Reward.}
\begin{table*}[]
\centering
\caption{\label{main_results} Evaluation results of different multimodal large language models (MLLMs) on \textit{ViRectify} benchmark. $\dagger$ denotes models fine-tuned on our dataset. The best results are shown in \textbf{bold} and the second best results are with {\ul underline}.}
\resizebox{2.\columnwidth}{!}{
\begin{tabular}{lcccccccccccc}
\toprule[1pt]
\multicolumn{1}{l|}{\multirow{2}{*}{\textbf{Models}}} & \multicolumn{3}{c|}{\textbf{Dynamic Perception}}                      & \multicolumn{3}{c|}{\textbf{Scientific Reasoning}}                    & \multicolumn{3}{c|}{\textbf{Embodied Decision}}                         & \multicolumn{3}{c}{\textbf{Total}}               \\ \cmidrule{2-13} 
\multicolumn{1}{l|}{}                                 & Acc$_i$        & IoU            & \multicolumn{1}{c|}{Acc$_r$}        & Acc$_i$        & IoU            & \multicolumn{1}{c|}{Acc$_r$}        & Acc$_i$        & IoU            & \multicolumn{1}{c|}{Acc$_r$}        & Acc$_i$        & IoU            & Acc$_r$        \\ \midrule
\multicolumn{13}{c}{\textit{Proprietary Models}}                                                                                                                                                                                                                                                                                 \\ \midrule
\multicolumn{1}{l|}{Gemini-2.5-Flash}                 & 71.88          & 41.73          & \multicolumn{1}{c|}{35.73}          & 71.73          & 46.01          & \multicolumn{1}{c|}{24.50}    & 76.39          & 49.19          & \multicolumn{1}{c|}{29.36}          & 72.64          & 44.96          & 30.92    \\
\multicolumn{1}{l|}{Gemini-2.5-Pro}                 & 75.34          & {\ul 42.71}           & \multicolumn{1}{c|}{{\ul 36.86}}           & 75.03           & 47.16          & \multicolumn{1}{c|}{{\ul 24.93}}     & 79.15          & {\ul 52.73}           & \multicolumn{1}{c|}{{\ul 30.22}}          & 76.01           & 45.68           & {\ul 31.77}     \\
\multicolumn{1}{l|}{GPT-4o}                           & 74.25          & 43.29 & \multicolumn{1}{c|}{36.17} & 72.83          & {\ul 47.82} & \multicolumn{1}{c|}{24.32}          & 78.41          & 50.40    & \multicolumn{1}{c|}{29.43} & 75.73          & {\ul 46.15} & 31.33 \\
\multicolumn{1}{l|}{GPT-5}                           &77.51 	&\textbf{44.07} 	&\multicolumn{1}{c|}{\textbf{37.33}} 	&76.97 	&\textbf{48.25} 	&\multicolumn{1}{c|}{\textbf{25.16}}	  &80.32 	&\textbf{52.96} 	&\multicolumn{1}{c|}{\textbf{30.54}} 	&77.98 	&\textbf{46.93} 	&\textbf{31.94}  \\\midrule
\multicolumn{13}{c}{\textit{Open-Source Models}}                                                                                                                                                                                                                                                                                 \\ \midrule
\multicolumn{1}{l|}{InternVL3-2B}                     & 25.18          & 17.31          & \multicolumn{1}{c|}{8.07}           & 25.55          & 15.73          & \multicolumn{1}{c|}{6.15}           & 25.52          & 24.03          & \multicolumn{1}{c|}{8.34}           & 25.39          & 18.01          & 7.39           \\
\multicolumn{1}{l|}{InternVL3-8B}                     & 42.14          & 25.23          & \multicolumn{1}{c|}{15.87}          & 42.05          & 25.58          & \multicolumn{1}{c|}{11.57}          & 43.48          & 37.08          & \multicolumn{1}{c|}{18.74}          & 42.39          & 27.85          & 14.98          \\
\multicolumn{1}{l|}{InternVL3-14B}                    & 50.13          & 27.16          & \multicolumn{1}{c|}{15.97}          & 48.63          & 25.05          & \multicolumn{1}{c|}{12.07}          & 55.15          & 23.09          & \multicolumn{1}{c|}{19.22}          & 49.97          & 29.08          & 15.40          \\
\multicolumn{1}{l|}{Qwen2.5-VL-3B}                    & 29.20          & 24.07          & \multicolumn{1}{c|}{10.15}          & 31.29          & 24.66          & \multicolumn{1}{c|}{8.69}           & 35.39          & 38.83          & \multicolumn{1}{c|}{8.02}           & 31.20          & 27.34          & 9.21           \\
\multicolumn{1}{l|}{Qwen2.5-VL-7B}                    & 36.29          & 24.85          & \multicolumn{1}{c|}{18.19}          & 34.55          & 25.85          & \multicolumn{1}{c|}{10.62}          & 36.71          & 39.91          & \multicolumn{1}{c|}{10.88}          & 35.77          & 28.44          & 13.94          \\
\multicolumn{1}{l|}{Qwen2.5-VL-32B}                   & 47.83          & 28.58          & \multicolumn{1}{c|}{19.48}          & 46.56          & 28.05          & \multicolumn{1}{c|}{12.05}          & 51.23          & 40.75          & \multicolumn{1}{c|}{11.75}          & 48.18          & 30.43          & 15.54          \\
\multicolumn{1}{l|}{Qwen2.5-VL-72B}                   & 65.82          & 39.84          & \multicolumn{1}{c|}{27.40}          & 66.84          & 38.61          & \multicolumn{1}{c|}{20.57}          & 70.67          & 52.48           & \multicolumn{1}{c|}{20.41}          & 66.88          & 42.90          & 23.67          \\
\multicolumn{1}{l|}{MiMo-VL-7B-SFT}                   & 31.67          & 28.06          & \multicolumn{1}{c|}{14.10}          & 31.40          & 28.95          & \multicolumn{1}{c|}{9.15}           & 30.83          & 46.01          & \multicolumn{1}{c|}{10.36}          & 31.43          & 31.61          & 11.71          \\
\multicolumn{1}{l|}{MiMo-VL-7B-RL}                    & 32.35          & 29.12          & \multicolumn{1}{c|}{16.84}          & 32.31          & 31.55          & \multicolumn{1}{c|}{11.13}          & 30.59          & 47.49          & \multicolumn{1}{c|}{13.13}          & 32.01          & 33.33          & 14.31          \\
\multicolumn{1}{l|}{R1-Onevision-7B-RL}               & 32.85          & 28.88          & \multicolumn{1}{c|}{12.43}          & 30.52          & 25.80          & \multicolumn{1}{c|}{7.76}           & 30.24          & 46.46          & \multicolumn{1}{c|}{10.85}          & 31.50          & 31.21          & 10.45          \\
\multicolumn{1}{l|}{GLM-4.1V-9B-Thinking}             & 49.91          & 29.02          & \multicolumn{1}{c|}{21.24}          & 48.49          & 29.89          & \multicolumn{1}{c|}{16.37}          & 51.35          & 34.99          & \multicolumn{1}{c|}{19.37}          & 49.71          & 30.47          & 19.25          \\ \midrule
\multicolumn{1}{l|}{Qwen2.5-VL-7B$^\dagger$}                    & 70.26          & 29.17          & \multicolumn{1}{c|}{20.53}          & 70.19          & 33.16          & \multicolumn{1}{c|}{21.47}          & 73.52          & 31.26          & \multicolumn{1}{c|}{20.94}          & 70.51          & 29.11          & 20.65          \\
\multicolumn{1}{l|}{MiMo-VL-7B-SFT$^\dagger$}               & 71.52          & 34.88          & \multicolumn{1}{c|}{22.94}          & 71.38          & 37.57          & \multicolumn{1}{c|}{22.23}           & 74.69          & 33.48          & \multicolumn{1}{c|}{23.37}          & 71.44          & 34.82          & 22.62          \\
\multicolumn{1}{l|}{InternVL3.5-8B$^\dagger$}             & 73.11          & 38.05          & \multicolumn{1}{c|}{23.33}          & 74.62          & 36.71          & \multicolumn{1}{c|}{22.65}          & 74.54          & 38.07          & \multicolumn{1}{c|}{23.64}          & 73.29         & 37.40         & 22.91          \\ \midrule
\multicolumn{1}{l|}{Ours w/o $\mathcal{R}_{time}$}    & {\ul 83.11}    & 29.11          & \multicolumn{1}{c|}{27.16}          & {\ul 79.74}    & 28.71          & \multicolumn{1}{c|}{20.01}          & {\ul 83.15}    & 33.42          & \multicolumn{1}{c|}{20.27}          & {\ul 81.78}    & 29.27          & 23.06          \\
\multicolumn{1}{l|}{Ours w/o $\mathcal{L}_{graph}$}   & 71.68          & 40.89          & \multicolumn{1}{c|}{29.33}          & 72.51          & 44.73          & \multicolumn{1}{c|}{21.72}          & 73.42          & 48.22          & \multicolumn{1}{c|}{27.10}          & 72.17          & 44.14          & 27.58          \\
\multicolumn{1}{l|}{Ours}                             & \textbf{83.84} & 42.24    & \multicolumn{1}{c|}{35.89}    & \textbf{80.78} &  46.05    & \multicolumn{1}{c|}{24.55} & \textbf{83.40} & 49.80          & \multicolumn{1}{c|}{28.76}    & \textbf{82.41} & 45.08    & 30.54          \\ \midrule[1pt]
\end{tabular}}
\end{table*}
Strict usage of \textless error\_identify\textgreater{}, \textless error\_graph\textgreater{}, \textless answer\textgreater{}, and \textless rationale\textgreater{} gives 1.0. Otherwise, the reward is 0.0.
The total reward signal integrates all three components with tunable weighting coefficients:
\begin{equation}
    \mathcal{R}_{\text{total}} = \lambda_1 \mathcal{R}_{\text{time}} + \lambda_2 \mathcal{R}_{\text{ans}} + \lambda_3 \mathcal{R}_{\text{fmt}}
\end{equation}
where $\lambda_1$, $\lambda_2$, and $\lambda_3$ control the relative importance of each reward component.

\section{Experiments}
We perform a systematic evaluation on \textit{ViRectify} across a range of state-of-the-art MLLMs. This section outlines our experimental configuration and key findings, organized into three parts: the evaluated models, experiment settings and metrics, and main results with analysis.
\subsection{Evaluated Models}
We evaluate 16 state-of-the-art MLLMs, including 12 open-source models ranging from 2B to 72B parameters and 4 proprietary models. The open-source models comprise Qwen2.5-VL \cite{qwen2.5-vl}, InternVL3 \cite{internvl3}, InternVL3.5 \cite{internvl3.5}, MiMo-VL \cite{mimo-vl}, R1-OneVision-7B-RL \cite{r1-onevision}, and GLM-4.1V-9B-Thinking \cite{glm4.1v}. The proprietary models include GPT-4o, GPT-5 \cite{gpt4}, Gemini-2.5-Flash and Gemini-2.5-Pro \cite{gemini2.5}. Model details are provided in Appendix.
\subsection{Experiment Settings and Metrics}

\noindent \textbf{Implementation Details.}
All experiments are conducted on 8 NVIDIA A100-80G GPUs. We use Qwen2.5-VL-7B as the backbone of our trajectory evidence-driven correction framework. For step-wise error trajectory modeling, trainable LoRA layers with rank 16 are added to all linear layers, trained for 3 epochs with a learning rate of 1e-5 and a maximum sequence length of 30,000. For evidence-guided correction, we generate 8 samples per input using top-p sampling (temperature 1.0, top-p 0.9). The KL divergence coefficient is set to 0.04, with a learning rate of 2e-6. The reward function weights are $\lambda_1 = 0.5$, $\lambda_2 = 0.4$, and $\lambda_3 = 0.1$. 
Further evaluation details are in Appendix.

\noindent \textbf{Evaluation Metrics.}
We use three primary metrics to evaluate error corrections.
Error Identification Accuracy (Acc$_{i}$) measures the average step-wise accuracy of identifying erroneous reasoning steps by comparing predicted and ground-truth error labels.
IoU assesses the overlap between predicted and reference timestamps of the video-grounded rationale.
Rationale Accuracy (Acc$_{r}$) evaluates whether the model’s rationale leads to the correct final answer, using GPT-4o to automatically compare predicted and ground-truth answers, assigning 1 for a match and 0 otherwise.

\subsection{Main Results}

Table \ref{main_results} presents the overall evaluation results of various models on the \textit{ViRectify} benchmark. We summarize four key findings:
\textbf{(i) Model scale and reasoning capability are critical performance drivers}. 
Larger models achieve higher accuracy in both error identification and video-guided rationale. For instance, as Qwen2.5-VL scales from 3B to 72B, Acc$_{i}$ improves from 31.20 to 66.88 and Acc$_{r}$ from 9.21 to 23.67, with InternVL3 exhibiting a similar trend.
Models with stronger reasoning abilities perform better in both error identification and video-guided rationale. For example, MiMo-VL-7B RL consistently outperforms its SFT version across all metrics. Proprietary models significantly surpass open-source counterparts by over 5 percentage points on Acc$_{r}$. However, even these advanced models achieve only about 30\% accuracy in video-guided rationale, highlighting their limited capacity to effectively utilize critical video information for reasoning correction and underscoring the challenge of our benchmark.
\textbf{(ii) Domain-level results reveal a gap between perception and reasoning.}
Models show similar accuracy in error identification across domains but vary significantly in video-guided rationale performance. While nearly all models excel in the dynamic perception domain, their performance declines notably in the more challenging scientific reasoning and embodied decision-making domains. It suggests that current models possess strong visual perception capabilities, but still face substantial limitations in reasoning and decision-making.
\textbf{(iii) Our framework effectively enhances the model’s performance.} 
Proprietary Models generally exhibit superior performance on the IoU and Acc$_r$ metrics, owing to their strong multimodal reasoning ability. Nevertheless, our framework attains a peak error identification accuracy of 82.41 and matches leading proprietary models in video-guided rationale generation, confirming its effectiveness. Notably, in the demanding Scientific Reasoning domain, our model achieves the Acc$_{r}$ of 24.55, significantly outperforming the larger Qwen2.5-VL-72B. These findings demonstrate our framework’s strong capability in correcting reasoning errors. 

\subsection{Ablation Study}
Table \ref{main_results} also reports the ablation results, from which several insights emerge:
(i) Compared with the zero-shot setting, supervised fine-tuning (SFT) substantially enhances the model’s error identification and correction abilities. Acc$_r$ rises from 13.94 to 20.65, representing a relative improvement of nearly 100\% in Acc$_i$. This demonstrates that SFT provides the model with a solid foundation for recognizing erroneous solutions.
(ii) Removing the graph loss consistently degrades performance across all metrics, with Acc$_i$ dropping from 82.41 to 72.17. This decline highlights the importance of structured error modeling in capturing faulty reasoning steps and maintaining correction accuracy.
(iii) Excluding the timestamp-level alignment reward leads to a sharp reduction in temporal localization, as IoU decreases from 45.08 to 29.27, accompanied by a drop in Acc$_r$ from 30.54 to 23.06. This indicates a weakened ability to ground the model’s reasoning in the relevant video evidence.
Overall, these findings confirm the complementary benefits of graph-based error modeling and fine-grained temporal supervision in improving video-guided error correction.
\begin{figure}[!]
  \centering
  \includegraphics[scale=0.268]{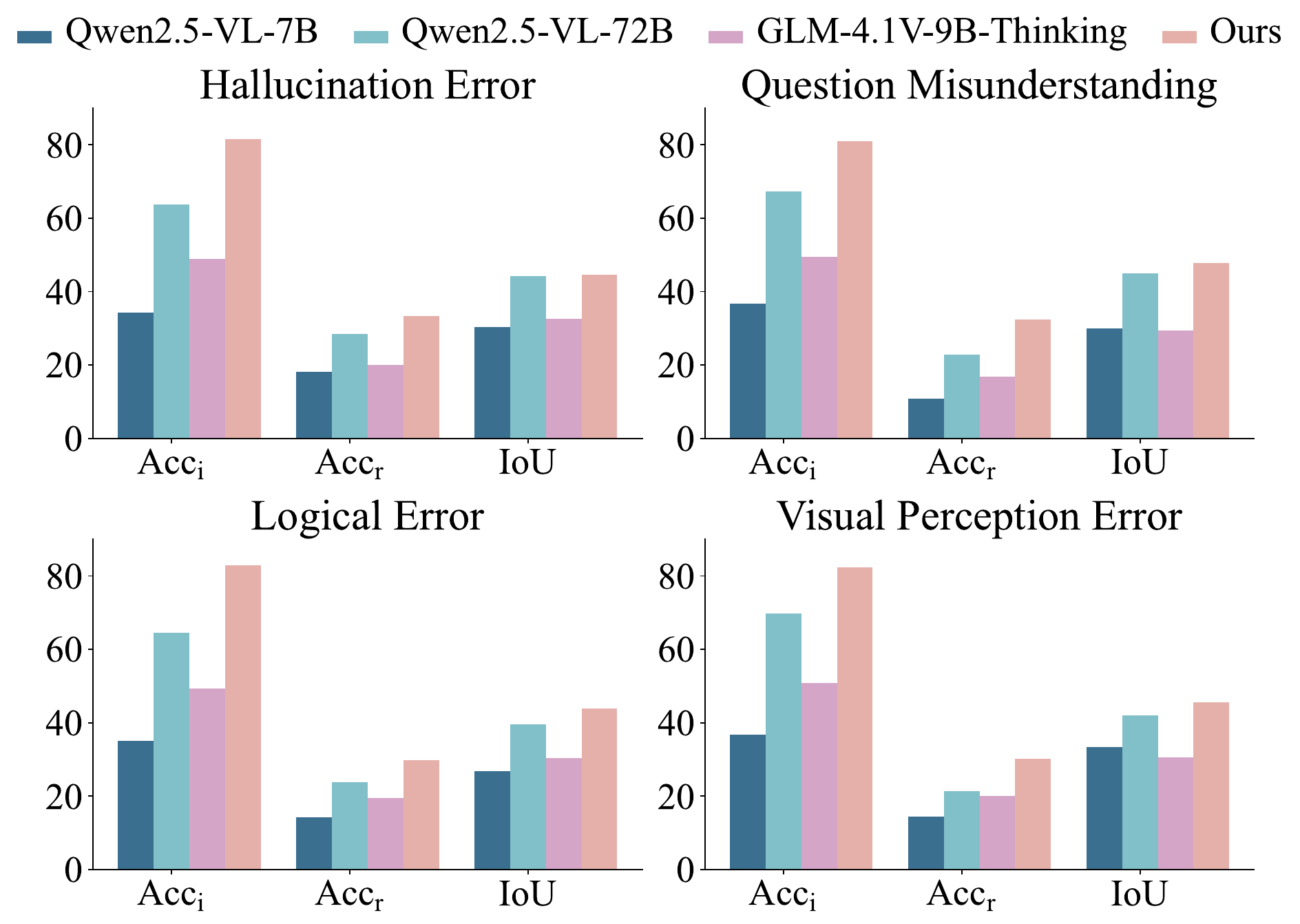}
  \caption{Performance across different error types.}
\label{err_analysis}
\end{figure}
\section{Analysis}

To gain deeper insights into the capabilities and limitations of current MLLMs and our framework, we conduct three investigations: (i) Error analysis across different models.
(ii) Cross-Model Evaluation.
(iii) Application on reflection learning to our benchmark.
\begin{figure}[!]
  \centering
  \includegraphics[scale=0.478]{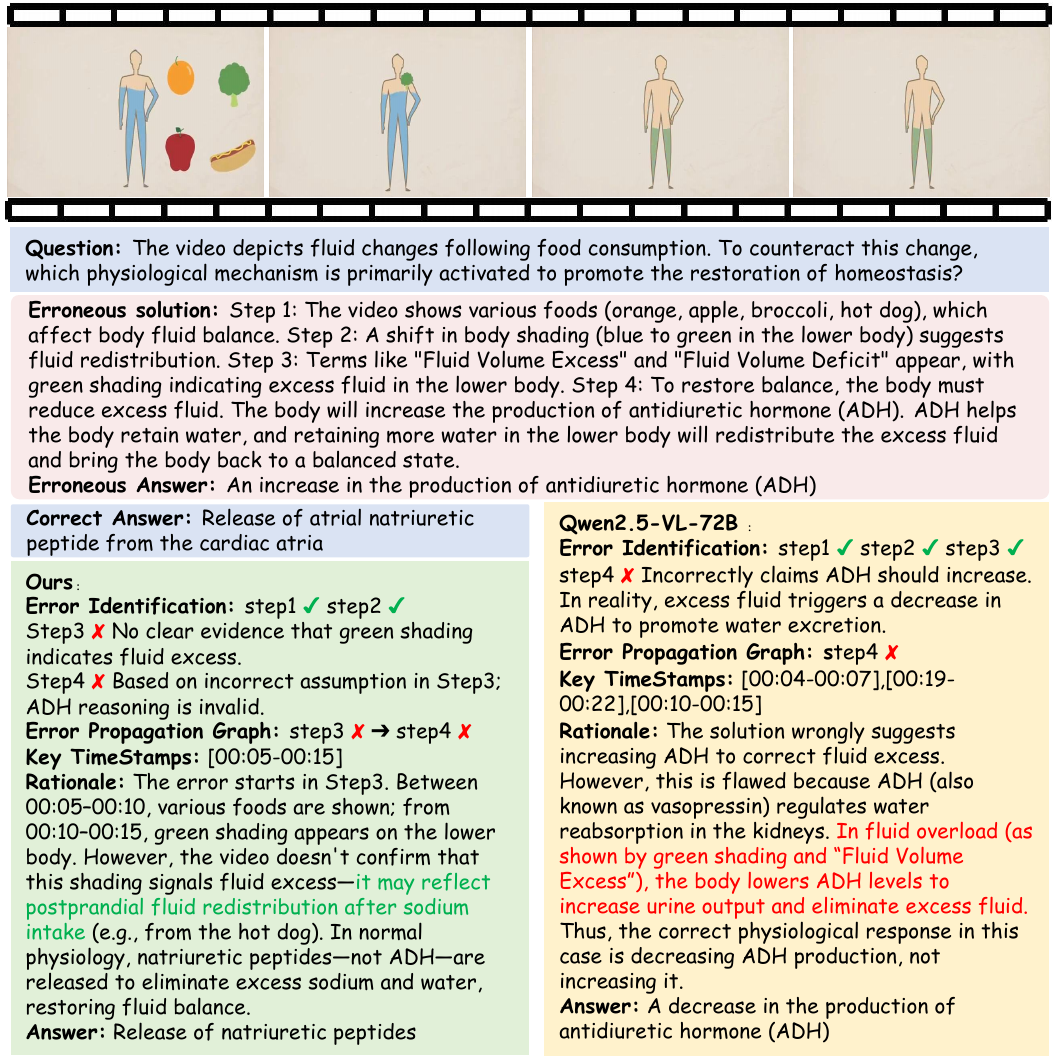}
  \caption{Case study 
  of Qwen2.5-VL-72B and our framework.}
\label{case_study}
\end{figure}
\subsection{Error Analysis}

\noindent \textbf{Analysis across Different Error Types.}
As shown in Fig. \ref{err_analysis}, visual perception and hallucination errors are easier to identify, while logical errors remain the most challenging for both identification and correction. Models with stronger reasoning abilities perform better in correcting logical errors. For example, GLM-4.1V-9B-Thinking exceeds Qwen2.5-VL-7B by 7\%, and Qwen2.5-VL-72B outperforms its 7B counterpart by 10\% in Acc$_r$. Our method consistently enhances error detection and correction across all error types, surpassing Qwen2.5-VL-72B. 

\noindent \textbf{Qualitative Comparison.}
Fig. \ref{case_study} presents a case where both Qwen2.5-VL-72B and our framework attempt to correct an erroneous solution. While the 72B model identifies a faulty reasoning step, it overlooks the error’s origin in a previous step and its propagation, leading to focus on irrelevant visual cues and an incorrect correction. In contrast, our model explicitly traces the error propagation and aligns the correction with pertinent video evidence, resulting in more accurate reasoning and correction.
More cases can be found in the Appendix.
\begin{figure}[!]
  \centering
  \includegraphics[scale=0.35]{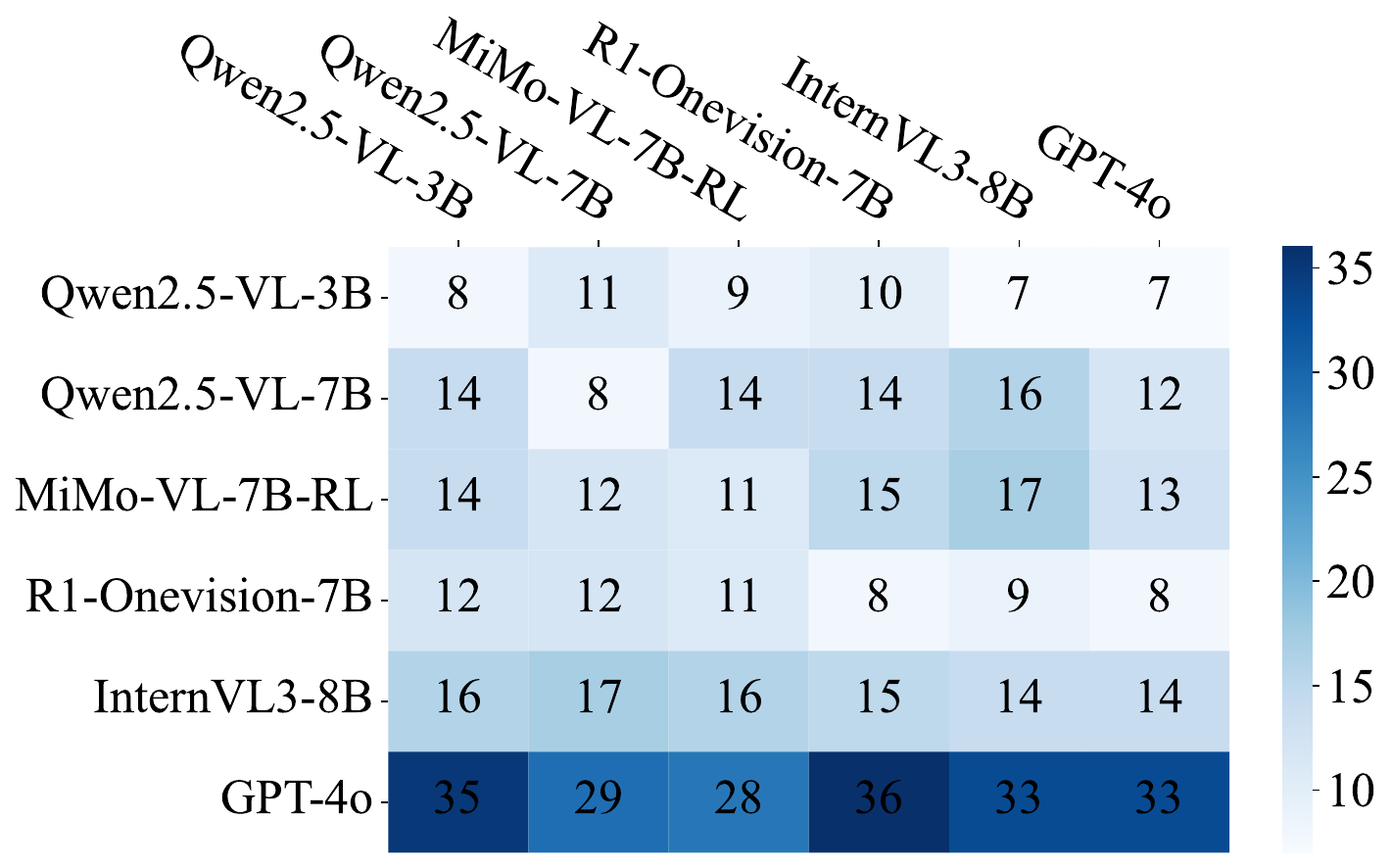}
  \caption{Cross-model performance comparison. Each cell ($x$, $y$) in the heatmap indicates the Acc$_r$ achieved when model $x$ acts as the evaluator for responses generated by model $y$.}
\label{cross_model}
\end{figure}
\subsection{Cross-Model Evalution}

Figure \ref{cross_model} presents the performance of different models, each acting as an evaluator to detect and correct reasoning errors in others’ outputs.
We observe two notable phenomena:
Smaller models exhibit overconfidence in their own erroneous solutions, showing limited ability to recognize or correct their own mistakes. In contrast, larger models demonstrate stronger error detection and rectification capabilities, indicating a more robust capacity to evaluate and refine their reasoning processes.
Moreover, larger models achieve higher accuracy when correcting the erroneous solutions of smaller models, whereas smaller models struggle to revise the outputs of larger ones effectively. For instance, GPT-4o attains 33\% accuracy when evaluating erroneous solutions from InternVL3-8B, while InternVL3-8B achieves only 14\% when assessing GPT-4o. This asymmetry highlights the superior reasoning capabilities of larger models, enabling them to better analyze and rectify the errors made by smaller models. In contrast, smaller models, constrained by their limited reasoning ability, often fail to accurately identify and rectify the more complex reasoning errors produced by larger models.
\begin{table}[!]
\centering
\caption{\label{cor_analysis} Effectiveness of error identification and rationale}
\resizebox{1.05 \columnwidth}{!}{
\begin{tabular}{l|cc}
\toprule[1pt]
                     & \textbf{Error Identification} & \textbf{Rationale} \\ \midrule
Ground Truth         & 27.08                         & 65.70                \\ \midrule
Qwen2.5-VL-7B       & 6.40                          & 7.69                 \\
Qwen2.5-VL-32B      & 6.76                          & 9.96                 \\
Qwen2.5-VL-72B      & 11.03                         & 14.95                \\
InternVL3-8B         & 6.25                          & 7.11                 \\
GLM-4.1V-9B-Thinking & 9.96                          & 16.02                \\
MiMo-VL-7B-RL        & 7.83                          & 9.61                 \\
Ours                 & \textbf{13.52}                         & \textbf{16.25}                \\ \midrule[1pt]
\end{tabular}}
\end{table}
\subsection{Application on Reflection}

To further assess the quality of generated error corrections, we separately provide the error identification and rationale with video evidence (excluding the final answer) as input to Qwen2.5-VL-7B, prompting it to revise its reasoning and produce a corrected answer. 
As shown in Table \ref{cor_analysis}, providing either error identification or rationale with visual context significantly improves answer accuracy. It demonstrates that explicitly incorporating error-focused reasoning and visual evidence enhances the model's correction ability. However, a substantial gap remains between model-generated and ground-truth rationales: with ground-truth rationales, the model achieves 65.7\% accuracy, but it drops to 10\% using rationales generated by other models. This gap underscores the current models' limitations in producing effective, informative rationales for reflection.


\section{Conclusion}
In this paper, we present \textit{ViRectify}, a benchmark for evaluating models' ability to identify and correct errors in video reasoning. 
We design an AI-assisted annotation pipeline with rigorous human verification, which contains over 30\textit{K} instances with step-wise error identification and video-grounded rationales.
Evaluation of 16 leading MLLMs reveals persistent challenges in error correction and underscores the effectiveness of our proposed trajectory evidence–driven correction framework.
We believe that
\textit{ViRectify} offers a concrete new pathway to evaluate MLLMs' ability in video reasoning.\mbox{}
{
    \small
    \bibliographystyle{ieeenat_fullname}
    \bibliography{main}
}

\clearpage
\setcounter{page}{1}
\maketitlesupplementary

\newcommand{\openai}{\raisebox{-1.5pt}{\includegraphics[height=1.05em]{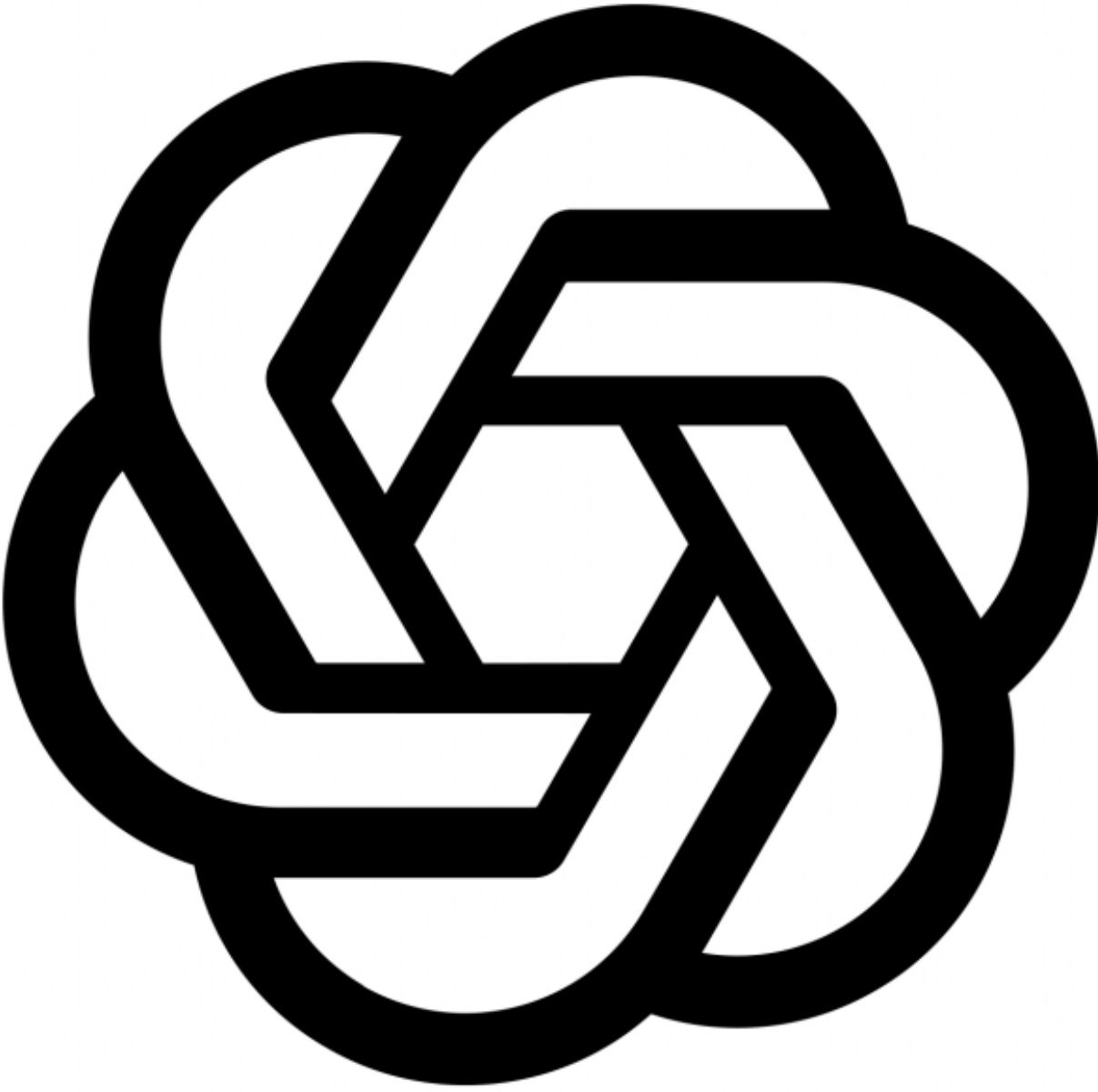}}\xspace}
\newcommand{\gemini}{\raisebox{-1.5pt}{\includegraphics[height=1.05em]{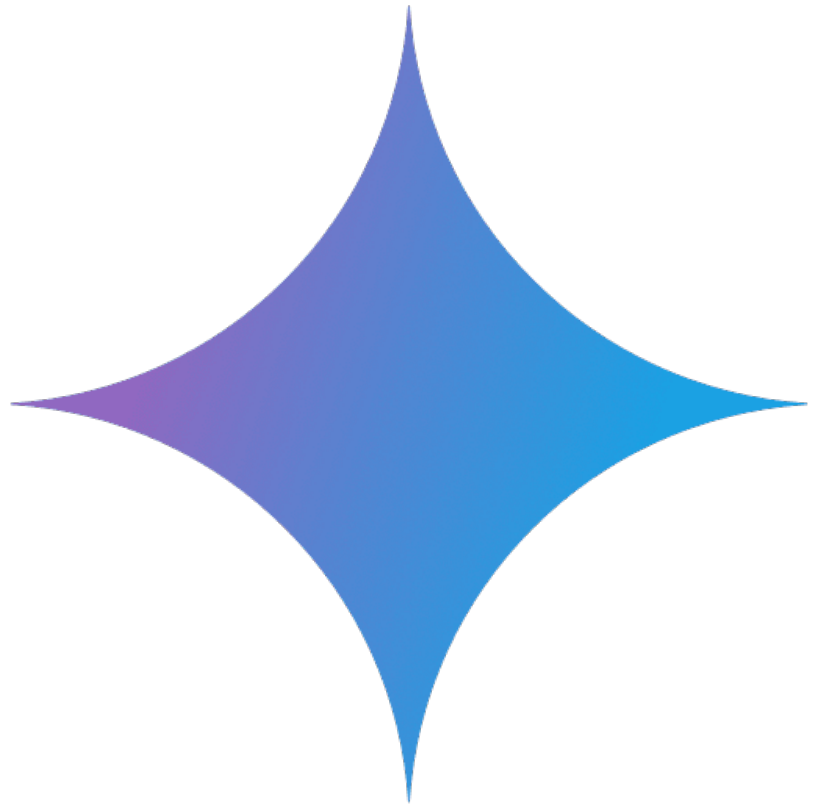}}\xspace}
\newcommand{\claude}{\raisebox{-1.5pt}{\includegraphics[height=1.05em]{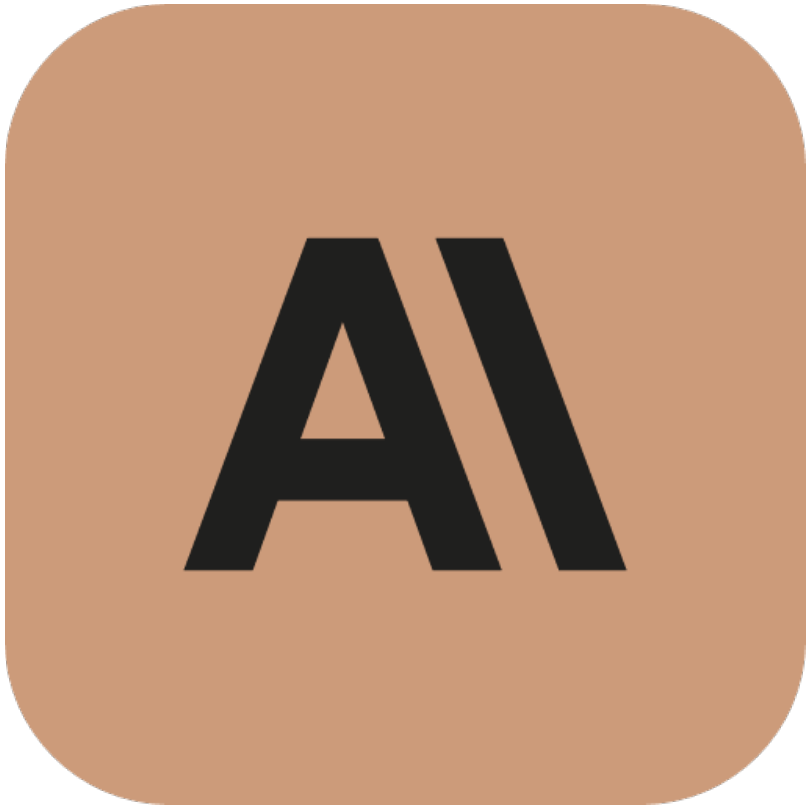}}\xspace}

\newcommand{\huggingface}{\raisebox{-1.5pt}{\includegraphics[height=1.05em]{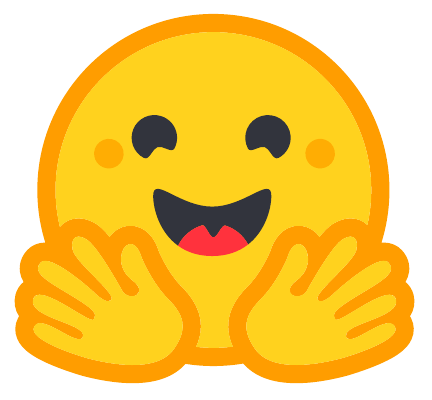}}\xspace}

\section{Benchmark Construction}
\subsection{Prompt for AI-assisted Annotation}
Table \ref{diff_sampling} presents the prompt used for difficulty-based rating. We retain only samples rated above 3 to ensure sufficient reasoning complexity. Qwen2.5-VL-32B is employed to generate captions using the prompt detailed in Table \ref{caption}. To assess the correctness of the answers in the video QA solutions with prompt in Table \ref{cot}, we adopt the evaluation prompt shown in Table \ref{eval}. We adopt the error injection prompt shown in Table \ref{err_inject}. Additionally, the prompts used for error identification and rationale annotation are provided in Table \ref{err_identify} and Table \ref{rationale}, respectively.

\subsection{Annotator Profile and Verification Protocol}
To ensure the diversity, challenge level, and correctness of the collected erroneous solutions and their corresponding corrections, we employ a three-stage human verification process involving three annotators with backgrounds in artificial intelligence, cognitive science, and multimodal reasoning. All annotators are graduate and senior undergraduate students with substantial experience in analyzing reasoning chains generated by large language models and prior involvement in related annotation tasks. Annotators were compensated at a standardized research rate of 10 USD/hour, reflecting the complexity of the verification process. Additional bonuses were offered for high-throughput and consistent performance across tasks.

\subsection{Error Corrections with Human Verification}
To ensure the quality and reliability of our annotated error correction data, we employ a three-stage human verification process: (i) We first assess the validity of the error tracing by examining the step-wise correctness labels, the accuracy of the provided error explanations, and the plausibility of the error propagation graph. (ii) We then verify the correction component by evaluating whether the selected key frames indeed contain critical visual cues relevant to error correction, and whether the revised solutions appropriately rectify the identified mistakes. (iii) Finally, we assess the overall consistency between the error tracing and the proposed correction to ensure logical coherence and alignment.

\section{Experimental Details}
\subsection{Model and Evaluation Details}
Table \ref{tab_model_detail} shows the details of the LMMs used in our experiments, including the corresponding LLMs and the visual encoders. 
All experiments are conducted with eight NVIDIA A100-80GB GPUs. 
Open-source models are deployed locally using vLLM, utilizing official checkpoints for model serving. 
Due to the input limitations of the proprietary model, we adopt a uniform sampling strategy to extract 32 evenly spaced frames from each video, which are then used as the model’s visual input.
The instruction templates used for training are provided in Table \ref{train_template}. Each video is represented by a sequence of frames with associated temporal annotations.
During evaluation, we use the default prompt and greedy decoding.

\subsection{Details about GRPO}
To identify samples suitable for error correction analysis, we conduct 8 independent rollouts per question using the first-stage model. We retain questions for which the model produces the correct answer in at least 1 but no more than 4 rollouts.
We employ Group Relative Policy Optimization (GRPO) to further optimize the fine-tuned models(\ie, Qwen2.5-VL-7B). The optimization is conducted with LoRA rank set to 32, a learning rate of 1e-5, a maximum sequence length of 30000, and a batch size of 1. To accelerate the GRPO inference process, we deploy the models using vLLM.

\subsection{More Experiments on Self-Correction}
To bridge the gap between model-generated and ground-truth rationales, we fine-tuned the Qwen2.5-VL-7B model using our previously proposed trajectory evidence-driven correction framework. Specifically, we performed rollouts on the training set and selected rationales that led to correct answers as supervision for self-correction training. This approach resulted in a substantial improvement in model performance. These findings suggest that our dataset can be effectively leveraged to train a dedicated model for generating self-correction data. By simply applying rejection sampling based on whether the answer label inferred from the generated rationale is correct, high-quality data for self-correction can be automatically obtained.

\section{More Cases}
\begin{figure*}[]
  \centering
  \includegraphics[scale=0.8]{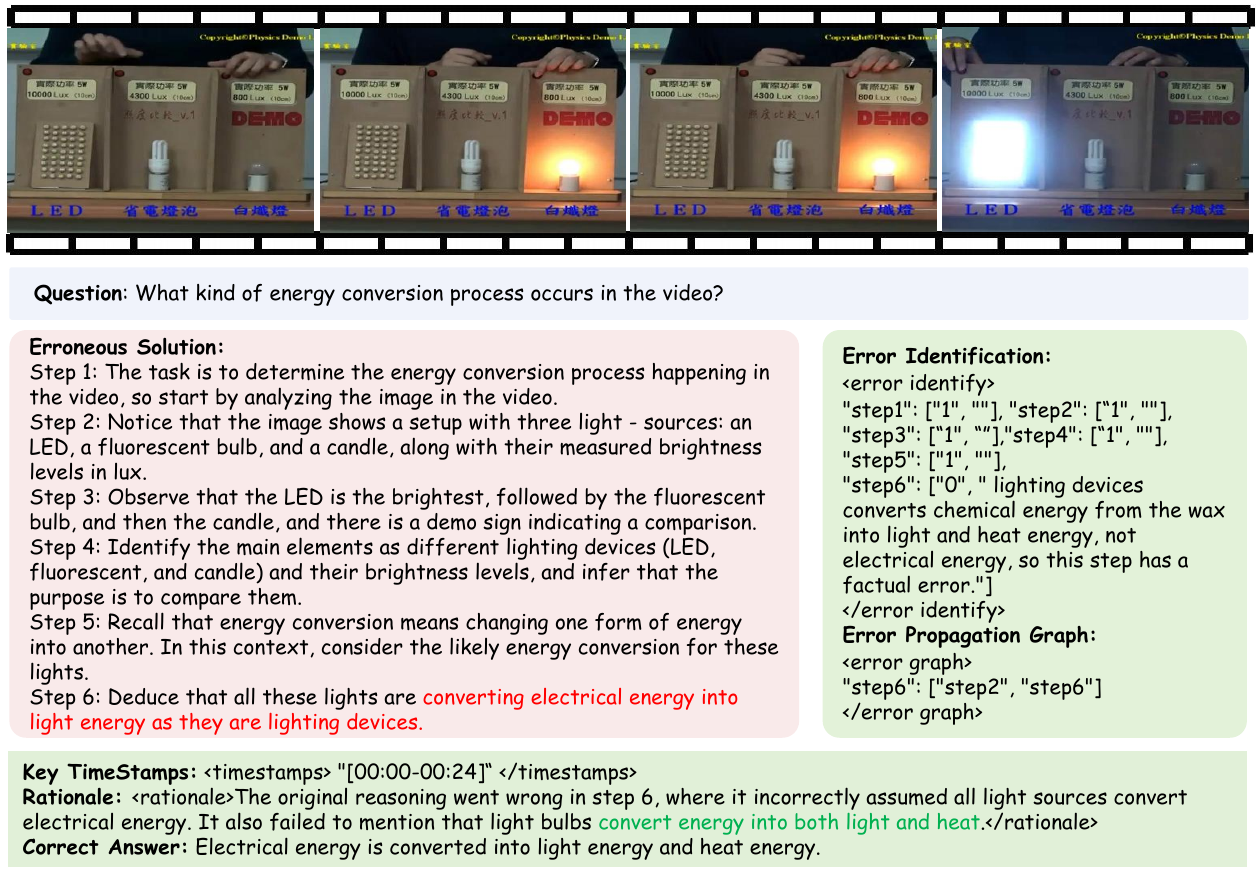}
  \caption{An example of scientific reasoning.} 
  \label{c_1}
\end{figure*}

\begin{figure*}[]
  \centering
  \includegraphics[scale=0.8]{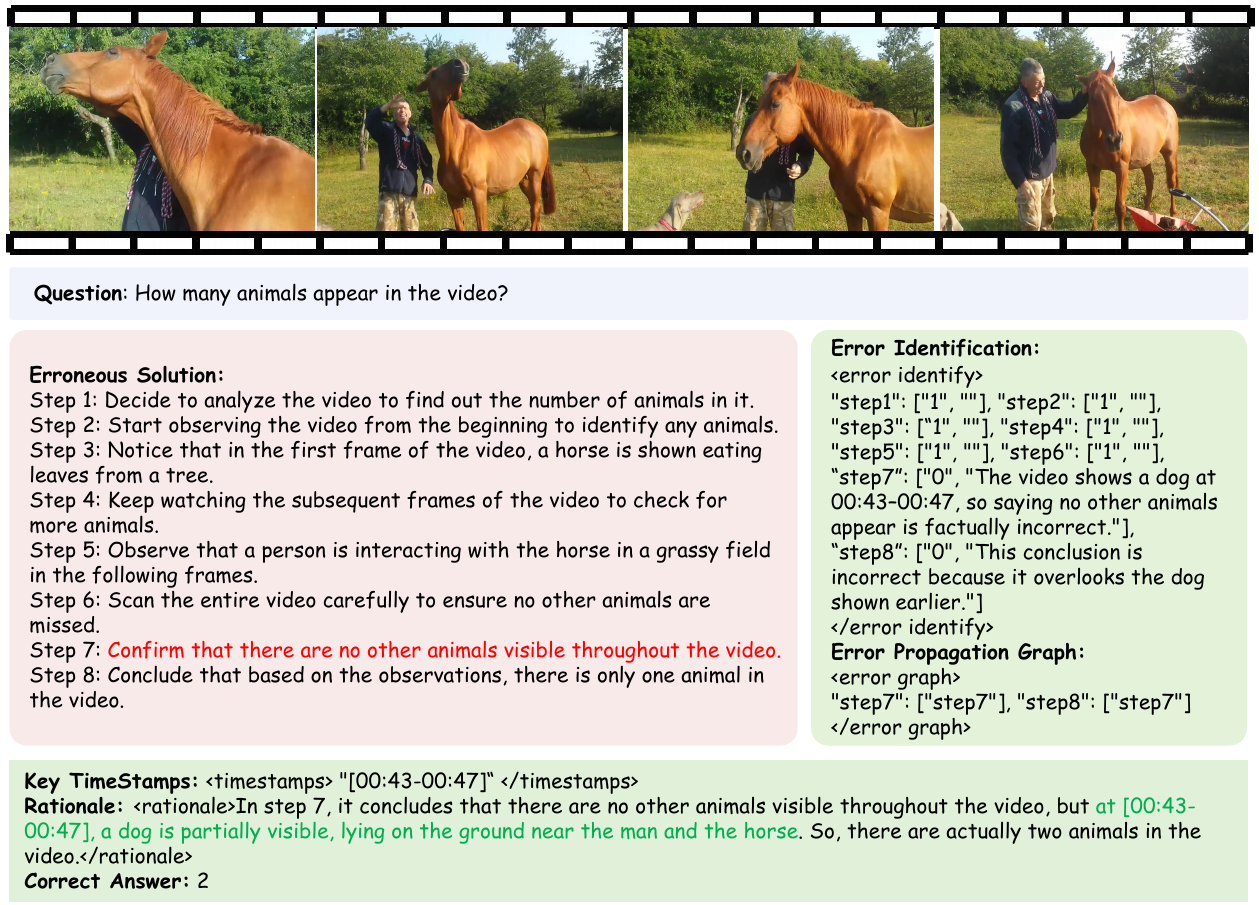}
  \caption{An example of scientific reasoning.} 
  \label{c_2}
\end{figure*}

\begin{figure*}[]
  \centering
  \includegraphics[scale=0.8]{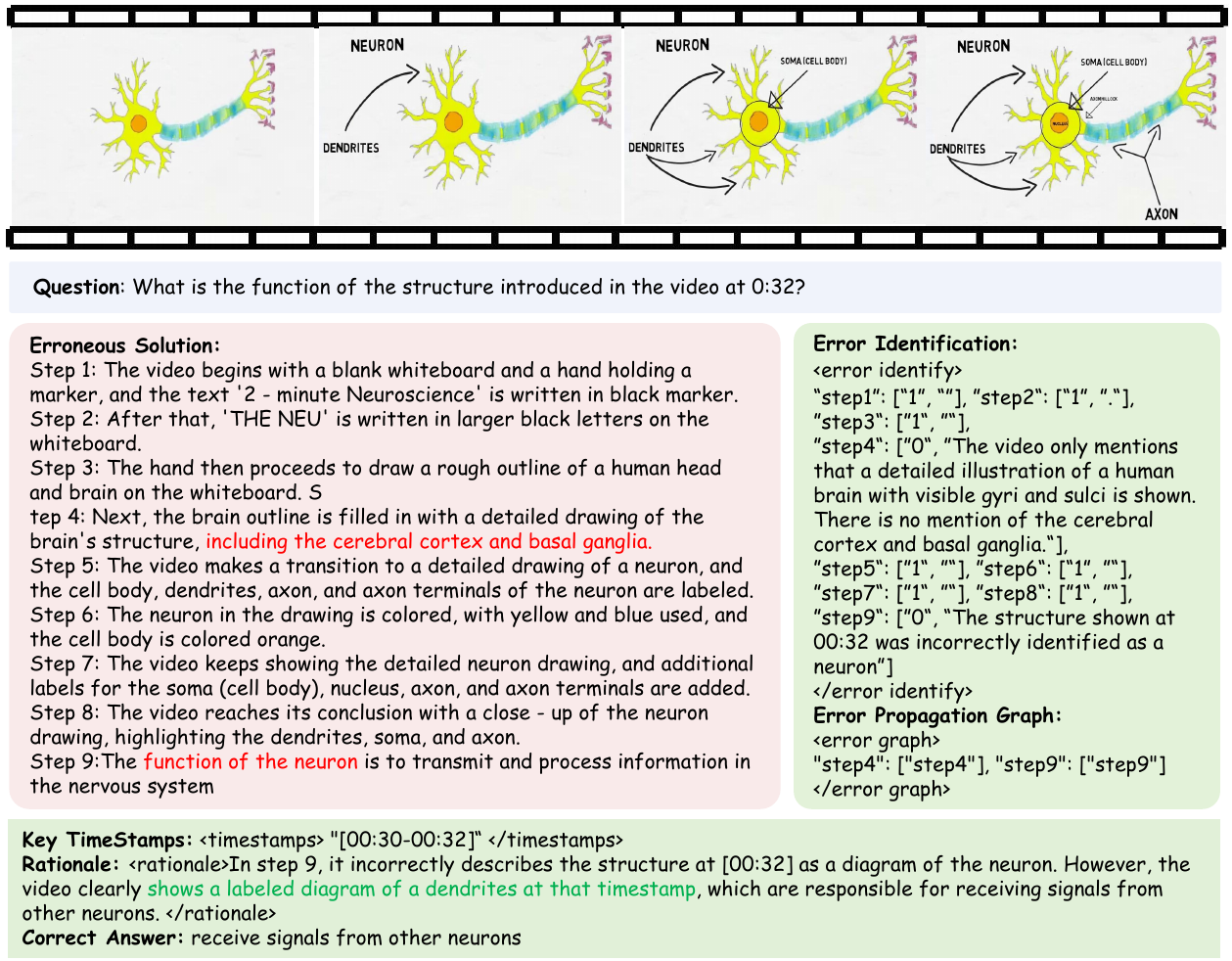}
  \caption{An example of scientific reasoning.} 
  \label{c_3}
\end{figure*}

\begin{figure*}[]
  \centering
  \includegraphics[scale=0.8]{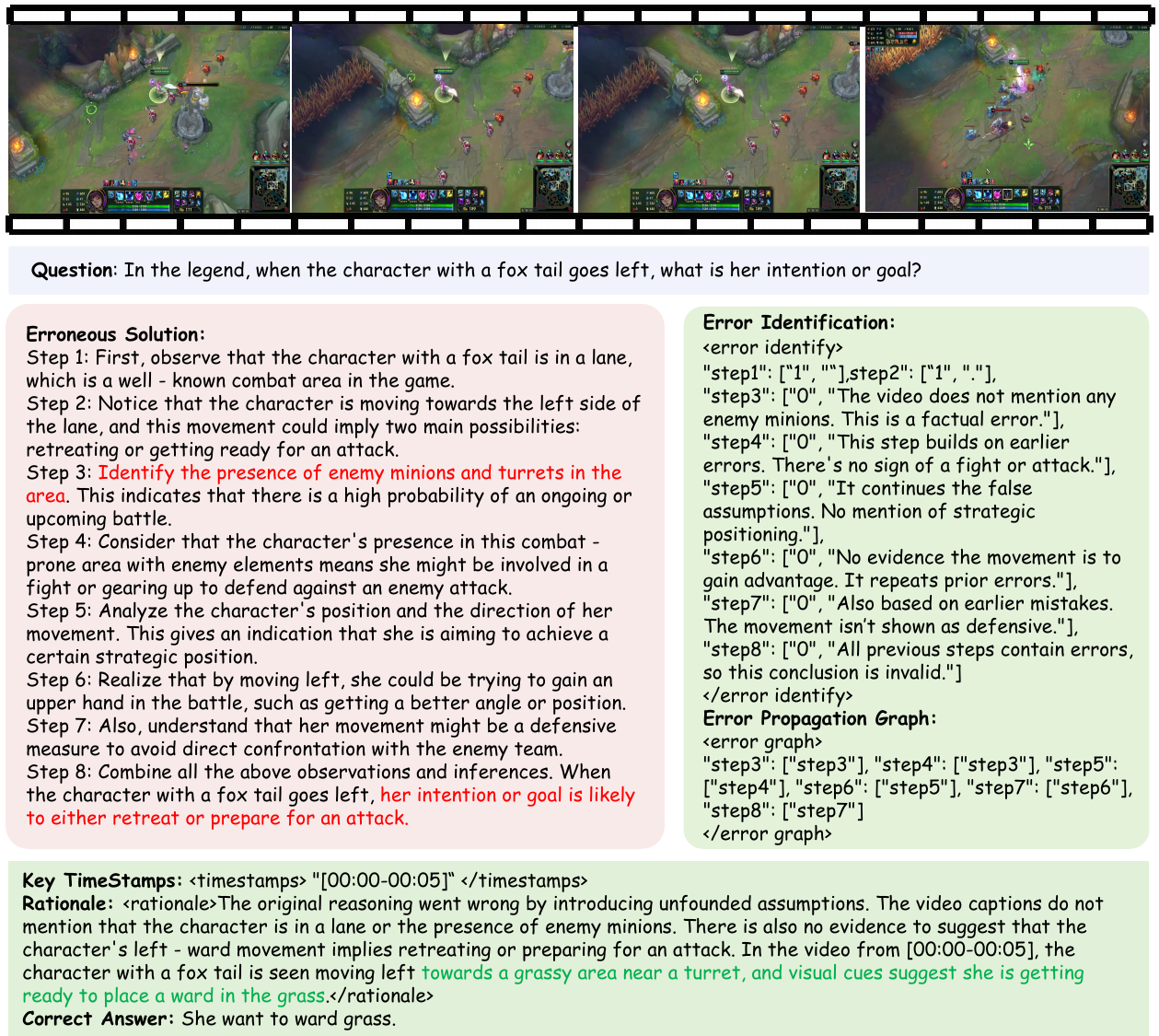}
  \caption{An example of dynamic perception.} 
  \label{c_4}
\end{figure*}

More cases across different domains are shown in Fig. \ref{c_1}, \ref{c_2}, \ref{c_3}, \ref{c_4} and \ref{c_5}

\begin{table*}[!]
    \centering
    \small
    \begin{spacing}{1.05}
    \caption{\label{diff_sampling}The prompt for difficuty-based sampling.}
    \resizebox{\textwidth}{!}{
    \begin{tabular}{p{\linewidth}}
    \toprule[1pt]
    Rate the following video question on a scale from 1 to 5 based on its reasoning complexity: \\
    1 – Answerable by simple surface-level observation \\
    2 – Requires basic video understanding (e.g., recognizing objects, actions, characters) \\
    3 – Requires connecting information across multiple segments \\
    4 – Involves deep reasoning or inference of implicit content    \\
    5 – Requires integrating video content with external knowledge and creative problem-solving \\
    Only output a single integer score (1–5), with no additional text. \\
    \midrule[1pt]
    \end{tabular}
    }
    \end{spacing}
\end{table*}

\begin{table*}[!]
    \centering
    \small
    \begin{spacing}{1.05}
    \caption{\label{eval}The prompt for evaluating the correctness of answers.}
    \resizebox{\textwidth}{!}{
    \begin{tabular}{p{\linewidth}}
    \toprule[1pt]
    You are a strict answer evaluator. Your task is to compare a ground truth answer and a predicted answer based on their semantic similarity and relevance to the question. \\
If the predicted answer accurately or closely matches the ground truth answer and answers the question well, output 1. Otherwise, output 0.\\
Only output 0 or 1. Do not provide any explanation. \\
    \midrule[1pt]
    \end{tabular}
    }
    \end{spacing}
\end{table*}

\begin{table*}[!]
    \centering
    \small
    \begin{spacing}{1.05}
    \caption{\label{caption}The prompt for video caption}
    \resizebox{\textwidth}{!}{
    \begin{tabular}{p{\linewidth}}
    \toprule[1pt]
    You are a video content analysis expert. Generate objective visual descriptions for short video clips. The Q\&A context is provided only for informational background and must NOT influence the caption content. \\
Input Information: \\
1. Global Q\&A Context: \\
 - Question: Insert question about the full video \\
 - Answer: Insert corresponding answer \\
2. Current Clip Details: \\
 - Visual Elements: Key objects/people/actions/scenes \\

Output Requirements: \\
1. Generate English caption in the format: \\
\texttt{<caption>} detailed visual description \texttt{</caption>} \\
2. The caption must: \\
 - Focus on observable visual details (objects, actions, scenes). \\
 - Use concrete, specific language (avoid vague terms like "someone" or "something"). \\
 - Remain neutral and factual—no interpretations or assumptions. \\
 - Describe clearly in one to three sentences based on the content of the video \\
3. The Q\&A context is only for reference; do not force a connection. \\
4. Do not inferences about purpose, meaning or context or assumptions about subjects' thoughts/intentions \\

Examples:   \\
\texttt{<caption>} A researcher in a white lab coat carefully transfers blue liquid between test tubes using a pipette.\texttt{</caption>} \\
    \midrule[1pt]
    \end{tabular}
    }
    \end{spacing}
\end{table*}

\begin{table*}[!]
    \centering
    \small
    \begin{spacing}{1.05}
    \caption{\label{cot}Prompt for Chain-of-Thought Reasoning in Video Question Answering.}
    \resizebox{\textwidth}{!}{
    \begin{tabular}{p{\linewidth}}
    \toprule[1pt]
    You are a step-by-step reasoning assistant for video question answering tasks.\\
Your task:\\
Given a video and a related question, carefully analyze the visual content and reason step-by-step before arriving at your final answer. Base your reasoning strictly on the video content, and follow the output format exactly.\\
Input Information: \\
1. Question: Insert question about the full video \\
2. Visual information: The sampling frames from the full video \\ 
You must follow the output format strictly! \\
Output Format: \\
\texttt{<think>}
Step 1: ...
Step 2: ...
...
Step N: ...
\texttt{</think>}
\texttt{<answer>}Your final answer here \texttt{</answer>}\\
    \midrule[1pt]
    \end{tabular}
    }
    \end{spacing}
\end{table*}

\begin{table*}[!]
    \centering
    \small
    \begin{spacing}{1.05}
    \caption{\label{err_inject}The prompt for injecting errors with proprietary model.}
    \resizebox{\textwidth}{!}{
    \begin{tabular}{p{\linewidth}}
    \toprule[1pt]
    You are an expert educator skilled in problem-solving and simulating authentic error chains in video-based learning. \\

Instructions:\\
1. Understand the following error categories and incorporate one or more of the following error types:\\
a. Logical Error: Logical flaws, invalid deductions, overgeneralizations  \\
b. Visual Perception Error: Object misidentification, overlooked details, spatiotemporal errors   \\
c. Question Misunderstanding: Misinterpreting the query's intent  \\
d. Hallucination Error: Adding non-existent elements/events \\ 

2. Task Requirements \& Constructing Error Chains\\
Your output should be a multi-step solution that leads to an incorrect final answer, but appears internally coherent and plausible.\\
You must inject errors subtly—they should not be explicitly acknowledged or flagged as incorrect. The chain should reflect the type of natural mistakes learners often make unknowingly.\\
a. Error Initiation: Introduce the first error at an early or intermediate step.\\
b. Error Propagation: Let this error logically influence at least two subsequent steps.\\
c. Error Termination: The final answer must be consistent with the reasoning chain but ultimately incorrect. \\
The overall logic should maintain surface-level plausibility to a casual observer, mirroring authentic reasoning failures.\\

3. Strict Constraints on Style and Reasoning \\
To maintain authenticity and pedagogical value: \\
a. No explicit signaling of errors (e.g., "wrongly infer…" or "assume…")\\
b. No unjustified eliminations (e.g., "It can’t be X, so it must be Y")\\
c. No meta-comments or hedging (e.g., "I'm not sure…", "There could be a mistake here…")\\
d. Do not reference any non-visual cues (e.g., captions, text, or audio descriptions) — all reasoning must emerge from visual analysis only\\
e. All observations and conclusions must appear confident and self-contained, as if from a well-meaning learner unaware of the embedded errors\\

4. Output Format:\\
\texttt{<error\_type>}Error Type\texttt{</error\_type>}
\texttt{<think>}
Step 1: ...
Step 2: ...
...
Step N: ...
\texttt{</think>}
\texttt{<answer>}Your final answer here \texttt{</answer>}\\
    \midrule[1pt]
    \end{tabular}
    }
    \end{spacing}
\end{table*}

\begin{table*}[!]
    \centering
    \small
    \begin{spacing}{1.05}
    \caption{ \label{err_identify} The prompt for error identification annotation.}
    \resizebox{\textwidth}{!}{
    \begin{tabular}{p{\linewidth}}
    \toprule[1pt]
You're an expert in reasoning analysis and video QA. \\
Input Information:\\
1. A video (represented by its caption segments, each with a timestamp)\\
2. A question about the video\\
3. Correct Chain of Thought (CoT)\\
4. Erroneous Chain of Thought (eCoT)\\
Task: Analyze the eCoT step-by-step to detect reasoning errors.\\
Error Detection Rules:\\
1. Correct Step:
Mark as ['1', ''] if:\\
 - Factually accurate (matches video captions)\\
 - Logically sound (valid inference)\\
 - Independent of prior errors \\
2. Incorrect Step:
Mark as ['0', 'explanation'] if:\\
 - Contains factual error (contradicts video)\\
 - Has logical flaw (invalid inference)\\
 - Propagates prior error (even if reasoning seems sound)\\
 - Introduces new error\\
Explanation must be specific: State what's wrong in this step (don't say "error passed on")\\

Error Propagation Graph Rules\\
1. Keys: Only include steps marked as incorrect ('0' in \texttt{<error\_identify>})\\
2. Values: Arrays specifying direct error sources\\
Output Format:\\
\texttt{<error\_identify>}\{ 'step1': ['1', ''], 'step2': ['0', 'error explanation'],\} \texttt{</error\_identify>}
\texttt{<error\_graph>}\{ 'step1': ['step1'], 'step2': ['step1'], 'step3': ['step2'] \}\texttt{</error\_graph>}\\

    \midrule[1pt]
    \end{tabular}}
    \end{spacing}
\end{table*}

\begin{table*}[!]
    \centering
    \small
    \begin{spacing}{1.05}
    \caption{ \label{rationale} The prompt for rationale annotation.}
    \resizebox{\textwidth}{!}{
    \begin{tabular}{p{\linewidth}}
    \toprule[1pt]
You are an expert in reasoning analysis and video question answering. \\
Input Information:\\
1. A video (represented by its caption segments, each with a timestamp),\\
2. A question about the video,\\
3. A correct Chain of Thought (CoT),\\
4. An erroneous Chain of Thought (eCoT) generated by a student or AI model,\\
5. An error identification for the eCOT.\\
Your task is to analyze the eCoT, and provide an overall correction using timestamp-aligned visual evidence from the video captions.\\

1. Timestamps:\\
 - List merged timestamp ranges (e.g., [00:12-00:20]) for most critical video evidence clips that directly correct the reasoning errors, and no more than three clips.\\
 - Combine adjacent clips contributing to the same correction (e.g., [00:11-00:12],[00:12-00:15] to [00:11-00:15])\\

2. Rationale:\\
 - Use natural language to describe how the original reasoning went wrong, referencing the key timestamps and what the visual content shows.\\
 - State the correct final answer to the question.\\
 - Never mention the captions or correct CoT or reasoning process.\\

Output Format:\\
\texttt{<timestamps>}[00:00-00:08],[...]\texttt{</timestamps>}
\texttt{<rationale>}
rationale with video evidence
\texttt{<answer>}The correct answer should be \textbackslash boxed\{xxx\}\texttt{</answer>}
\texttt{</rationale>}\\
    \midrule[1pt]
    \end{tabular}
    }
    \end{spacing}
\end{table*}

\begin{table*}[!]
    \centering
    \small
    \begin{spacing}{1.05}
    \caption{ \label{train_template} The instruction templates used for training.}
    \resizebox{\textwidth}{!}{
    \begin{tabular}{p{\linewidth}}
    \toprule[1pt]
    [1s]\texttt{<image>}[2s]\texttt{<image>} ... \texttt{<image>}
You are an expert in reasoning analysis and video question answering. Given the following information: \\
 - A video (represented by a set of timestamped frames),\\
 - A question about the video,\\
 - A solution for the question (with step-by-step reasoning),\\
Your task is to analyze the solution, detect reasoning errors step by step, and provide an overall correction using timestamp-aligned visual evidence from the video.  \\

Guidelines:\\
1. Error identification:\\
 - For each step in the solution (e.g., "step1", "step2"...), judge whether the reasoning is correct.\\
 - If correct: Mark as '1' and leave the second element as an empty string.\\
 - If incorrect: Mark as '0' and provide a brief and specific explanation of what is wrong in the step.\\
2. Error Propagation Graph:\\
 - For each step with an error, list direct sources of errors in that step.\\

3. Timestamps:\\
 - List merged timestamp ranges (e.g., [00:12-00:20]) for most critical video evidence clips that directly correct the reasoning errors.\\

4. Ratioanale:\\
 - Use natural language to describe how the original reasoning went wrong, referencing the specific timestamps and what the visual content shows. State the correct final answer to the question. \\

    \midrule[1pt]
    \end{tabular}
    }
    \end{spacing}
\end{table*}

\begin{table*}[t!]
    \centering
      \caption{\label{tab_model_detail}The details of the models used in our experiments. }
    \small 
    \resizebox{0.95\textwidth}{!}{
    \begin{tabular}{@{}l|cc@{}}
    \toprule 
        LMMs    & LLM (Size) & Vision Encoder\\
     \midrule
          \multicolumn{3}{c}{\emph{Open-Source Models}} \\ 
     \midrule
    InternVL3-2B & Qwen2.5 (1.5B) 	& InternViT-300M \\ 
    InternVL3-8B & Qwen2.5 (7B) 	& InternViT-300M \\ 
    InternVL3-14B & Qwen2.5 (14B) 	& InternViT-300M \\ 
    InternVL3.5-8B & Qwen3 (8B) 	& InternViT-300M\\ 
    MiMo-VL-7B-SFT & MiMo (7B) 	& ViT-bigG\\ 
    MiMo-VL-7B-RL & MiMo (7B) 	& ViT-bigG \\ 
    Qwen2.5-VL-3B & Qwen2.5 (3B)	& ViT-bigG \\ 
    Qwen2.5-VL-7B & Qwen2.5 (7B)	& ViT-bigG \\ 
    Qwen2.5-VL-32B & Qwen2.5 (32B)	& ViT-bigG \\ 
    Qwen2.5-VL-72B & Qwen2.5 (72B)	& ViT-bigG \\ 
    R1-Onevision-7B & Qwen2.5 (7B)	& ViT-bigG \\ 
    GLM-4.1V-9B-Thinking & GLM-4 (9B) 	& AIMv2-Huge\\ 
    \midrule 
     \multicolumn{3}{c}{\emph{Proprietary Models}} \\ 
        \midrule 
    Gemini-2.5-Flash & 	N / A& N / A \\ 
    Gemini-2.5-Pro & 	N / A& N / A  \\ 
    GPT-4o    &  N / A  & N / A\\ 
    GPT-5    &  N / A  & N / A \\ 
    \bottomrule
    \end{tabular}
    }
    \label{tab:model_specs}
\end{table*}

\begin{figure*}[]
  \centering
  \includegraphics[scale=0.8]{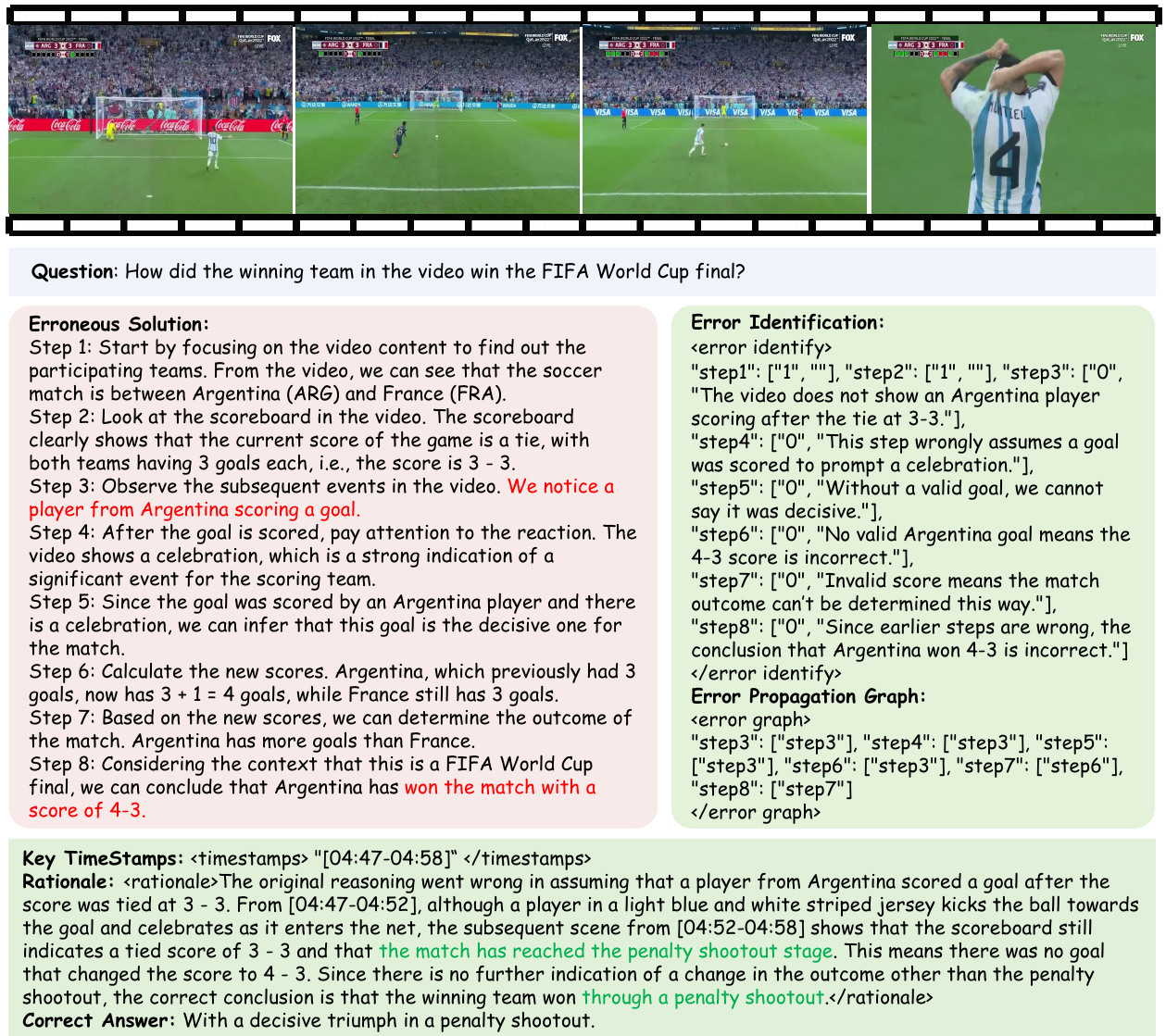}
  \caption{An example of dynamic perception.} 
  \label{c_5}
\end{figure*}
\begin{figure*}[]
  \centering
  \includegraphics[scale=0.8]{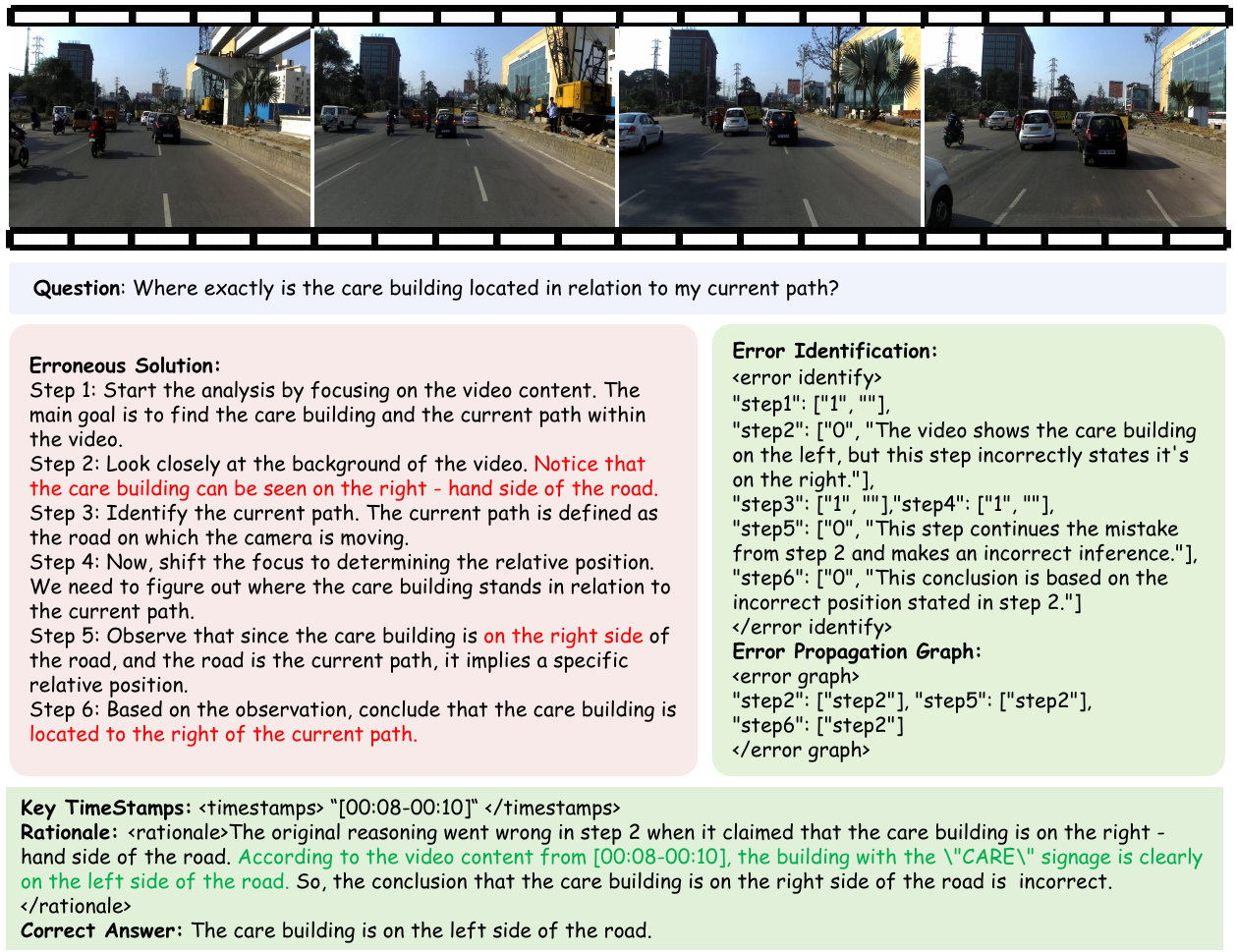}
  \caption{An example of embodied decision-making.} 
  \label{c_5}
\end{figure*}

\end{document}